\documentclass[10pt,journal,compsoc]{IEEEtran}

\ifCLASSOPTIONcompsoc
  \usepackage[nocompress]{cite}
\else
  \usepackage{cite}
\fi
\ifCLASSINFOpdf
  \usepackage[pdftex]{graphicx}
\else
\fi
\usepackage{amsmath}
\usepackage{amsfonts}
\usepackage{float}
\begin{document}
%
\title{InvKA: Gait Recognition via Invertible Koopman Autoencoder}
%
%
%
%

\author{Fan~Li, Dong~Liang, Jing~Lian, Qidong~Liu, Hegui~Zhu, Jizhao~Liu}

\markboth{Journal of \LaTeX\ Class Files,~Vol.~14, No.~8, August~2015}%
{Shell \MakeLowercase{\textit{et al.}}: Bare Advanced Demo of IEEEtran.cls for IEEE Biometrics Council Journals}
%



\IEEEtitleabstractindextext{%
\begin{abstract}
Most current gait recognition methods suffer from poor interpretability and high computational cost. To improve interpretability, we investigate gait features in the embedding space based on Koopman operator theory. The transition matrix in this space captures complex kinematic features of gait cycles, namely the Koopman operator. The diagonal elements of the operator matrix can represent the overall motion trend, providing a physically meaningful descriptor. To reduce the computational cost of our algorithm, we use a reversible autoencoder to reduce the model size and eliminate convolutional layers to compress its depth, resulting in fewer floating-point operations.  Experimental results on multiple datasets show that our method reduces computational cost to \textbf{1\%} compared to state-of-the-art methods while achieving competitive recognition accuracy (\textbf{98\%}) on non-occlusion datasets.
\end{abstract}

\begin{IEEEkeywords}
Gait recognition, Koopman operator, Invertible autoencoder
\end{IEEEkeywords}}

\maketitle

\IEEEdisplaynontitleabstractindextext

%
\IEEEpeerreviewmaketitle

\ifCLASSOPTIONcompsoc
\IEEEraisesectionheading{\section{Introduction}\label{sec:introduction}}
\else
\label{sec:introduction}
\fi

%
%
%
%

\IEEEPARstart{G}{ait} recognition involves authenticating or recognizing an individual's identity based on their walking posture or footprints \cite{zheng2022gait}. Previous approaches utilize deep learning techniques to extract and match features between frames in gait videos for identity authentication.\cite{hosni2020geometric,huang20213d,huang2021context,9351667,lin2021gait,zhu2023gait,wang2023dygait}. Gait recognition technology is essentially an identification algorithm. As such, the model must be allowed to be deployed on large-scale surveillance.\cite{alvarez2021exploring}.
It is widely recognized that \emph{mathematical theory-based methods are highly interpretable and elegant}, deep learning-based methods fit approximate solutions to complex algorithms quickly by associating features with labels\cite{GoodBengCour16}. Therefore, introducing a mathematical theoretical framework in gait recognition method is beneficial for simplifying computational processes.
\par
Here raise a question: \emph{Is there a method that combines both high interpretability and low computational cost}? Based on this idea, we have adopted \textbf{Koopman operator theory} and an \textbf{invertible autoencoder} based on neural networks to provide a simplified model with some interpretability and low computational cost for gait recognition, which is called the \textbf{InvKA}(Invertible Koopman Autoencoder) model. The Koopman operator theory provides a novel strategy for capturing temporal information from gait videos. The autoencoder solves the nonlinear transformation task in Koopman operator theory.
\par
Due to the extremely high computational complexity of convolutional layers(See Section \ref{Inertible}), the computational process becomes a black box lacking interpretability. Convolution on large 3D video data results in significant computational cost, limiting the deployment of this technology.
Most state-of-the-art methods have enormous model complexity (measured by FLOPs\footnote{FLOPs (Floating Point Operations) is a metric used to quantify the number of floating point operations executed during a computational task. It measures the number of floating point operations necessary for forward propagation, with a unit of 1.\cite{8622396}}and the number of model parameters).
The GaitSet\cite{9351667} model, for example, contains about 10 2D-convolutional layers, resulting in nearly \textbf{8.7G} FLOPs. Performing interpretability analysis on these floating-point operations is unrealistic. Additionally, the significant computational cost limits the model’s application in large-scale surveillance.
\par
To address these issues, our proposed approach for gait recognition exploits the periodic salience features of gait video streaming data to develop a simple and elegant descriptor(See Fig. \ref{fig:K}) that captures the geometric descriptions of various body parts. We achieve this by segmenting the long gait video stream into clips(See Section \ref{period analysis}) with gait cycles and using an invertible autoencoder(See Section \ref{Inertible}) to transform the periodic gait video into a low-dimensional embedding space. The embedding space exhibits the Markov property between successive frames of the original gait video(See Fig. \ref{fig:intro_full}. C), allowing the pose image of any frame can be determined solely from the previous frame and a fixed transition function (i.e., the Koopman operator). 
\par
Our method significantly reduces the video data volume by exploiting the optimal periodic kinematic representation of gait video stream data. Specifically, this approach reduces the computational cost by \textbf{2 orders} of magnitude for gait recognition while maintaining competitive recognition accuracy(\textbf{98\%}) on non-occlusion dataset(See Section \ref{validation}).
\par
Compared to existing models, our model has several outstanding attributes:

\begin{figure*}[!t]
\centering
\includegraphics[width=0.9\textwidth]{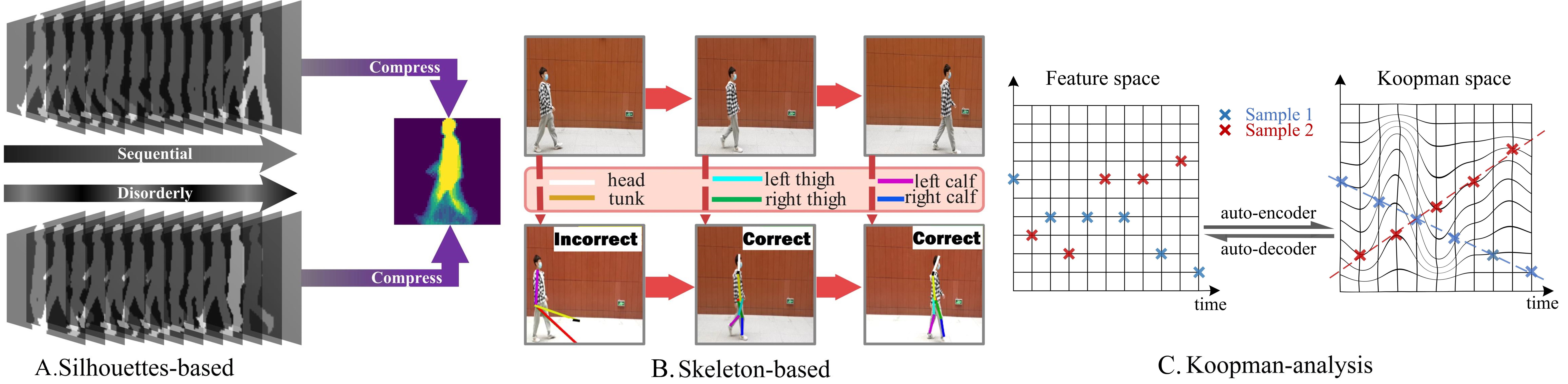}
\caption{\label{fig:intro_full}three images display different gait recognition methods: silhouettes-based (GEI), skeleton-based, and Koopman-based. GEI method ignores temporal information, and the skeleton-based method is prone to labeling errors. Koopman analysis method uses an autoencoder to transform the feature space and extract compressed features from video streaming data using a transition matrix with Markov Chain properties.}
\end{figure*}
\begin{itemize}
\item 
\textbf{Physically interpretable}: Our method extracts the kinematic feature of postures during the motion for classification by invertible autoencoder, which possesses a physically meaningful matrix.(See Section \ref{Koopman analysis}).
\item 
 \textbf{Low computational cost}: Our reversible autoencoder has improved  the structure of image encoder(See Section \ref{Inertible}). This has reduced the computational cost of our model to just one-hundredth of that of state-of-the-art models (See Section \ref{computional cost}).
\item 
\textbf{Capable of generating}: The Koopman operator contains nearly all kinematic information, enabling our model to generate synthetic gait frames through the decoder by utilizing the operator (See Appendix A).
\end{itemize}
\par
In summary, our proposed method for gait recognition leverages the unique periodic salience features of gait video data to develop an physically meaningful operator. To guarantee the \emph{decoder's generalization ability}, we employ an invertible autoencoder without convolutional layers. Our approach significantly reduces model complexity while preserving competitive recognition accuracy.
\section{Related works}
\par
Previous work includes silhouette-based gait recognition methods, skeleton-based gait recognition methods, and Koopman operator-based methods. 
\subsection{Silhouettes-based methods}
Silhouette-based gait recognition methods include \textbf{temporal-based methods} and \textbf{spatial-based methods}. These methods first view gait video stream data as a three-dimensional tensor, i.e., [W (width), H (height), L (length)]. Temporal-based methods primarily compress and adjust frames from the L dimension, focusing on the change process of postures. 
\par 
Temporal-based methods split data along the spatial dimension and primarily focus on the correlation between limbs and torso in gait. For instance, Fan et al. utilized Focal-convolution layers to divide each frame horizontally into multiple parts\cite{fan2020gaitpart}, perform convolutional operations on each part, and finally feed these extracted features into a fully connected layer. 
\par
Spatial-based methods preserve the sequential structure of the silhouettes, leveraging RNNs \cite{an2020performance}, 3D CNNs \cite{lin2020gait}, or GCNs \cite{li2020jointsgait} to improve feature extraction. However, these methods that utilize large-scale convolutions incur substantial FLOPs. 
Consequently, although spatial-based methods preserve time information in the data stream, the considerable computational cost of end-to-end methods limits their application.
\subsection{Skeleton-based methods}
Skeleton-based gait recognition methods begin by capturing the body joints using pose-estimation methods or depth cameras. Subsequently, the temporal relationships of joint positions are extracted using Recurrent Neural Networks (RNNs) \cite{liu2016memory} and Capsule Networks (CapsNets) \cite{chen2021receptor}. Finally, gait recognition is accomplished by matching joint features. Although skeleton-based methods are generally robust against appearance changes, they rely heavily on accurately detecting body joints, and training a model capable of precise human pose estimation is challenging. The difficulty arises from the fact that joint annotators are often hindered by occlusions caused by clothing, making it challenging to determine joint positions accurately. Clothing occlusions can cause even more significant errors when the model performs automatic annotation, resulting in lower gait recognition accuracy \cite{liu2022symmetry}. As shown in Fig. \ref{fig:intro_full}. B, when there is mutual occlusion between limbs and torso due to viewpoint, automatic annotators tend to make more errors in labeling. Recently, Liu et al. have proposed a method that removed redundant dependencies between joint features through graph convolution\cite{liu2020disentangling}. Combining spatial-based methods, using special convolution operators to directly extract features from skeleton graph sequences.

\subsection{Koopman operator-based methods}
The Koopman operator theory is a new technology for analyzing complex kinematic systems using dynamic mode decomposition (DMD), which enables the prediction of long-term behavior in observed data \cite{takeishi2017learning,liu2020hierarchical}. The theory employs a nonlinear mapping to project data from the spatiotemporal space onto the embedding space, where the resulting data stream exhibits Markovian properties. Specifically, the state of each temporal sample in the new space can be computed directly from the previous state and a fixed transition matrix without depending on earlier states, as illustrated in Fig.\ref{fig:intro_full}. C. This transition matrix is referred to as the Koopman operator. However, most existing Koopman operator research rarely focuses on representing kinematic features of periodic video streaming \cite{zhang2021cross}. Unlike existing models that use convolution-deconvolution structures for encoding and decoding, our proposed model divides the gait cycle to compress data and removes convolutional layers to simplify the model. By combining the reversible neural network \cite{dinh2014nice} and Koopman autoencoder, our proposed model guarantees that spatial transformations are surjective, enabling the creation of a non-loss autoencoder without using a convolution-deconvolution structure. This approach ensures the decoder's ability to generalize across the entire embedding space.
\section{Proposed Method}
\begin{figure*}[!htb]
\centering
\includegraphics[width=1\textwidth]{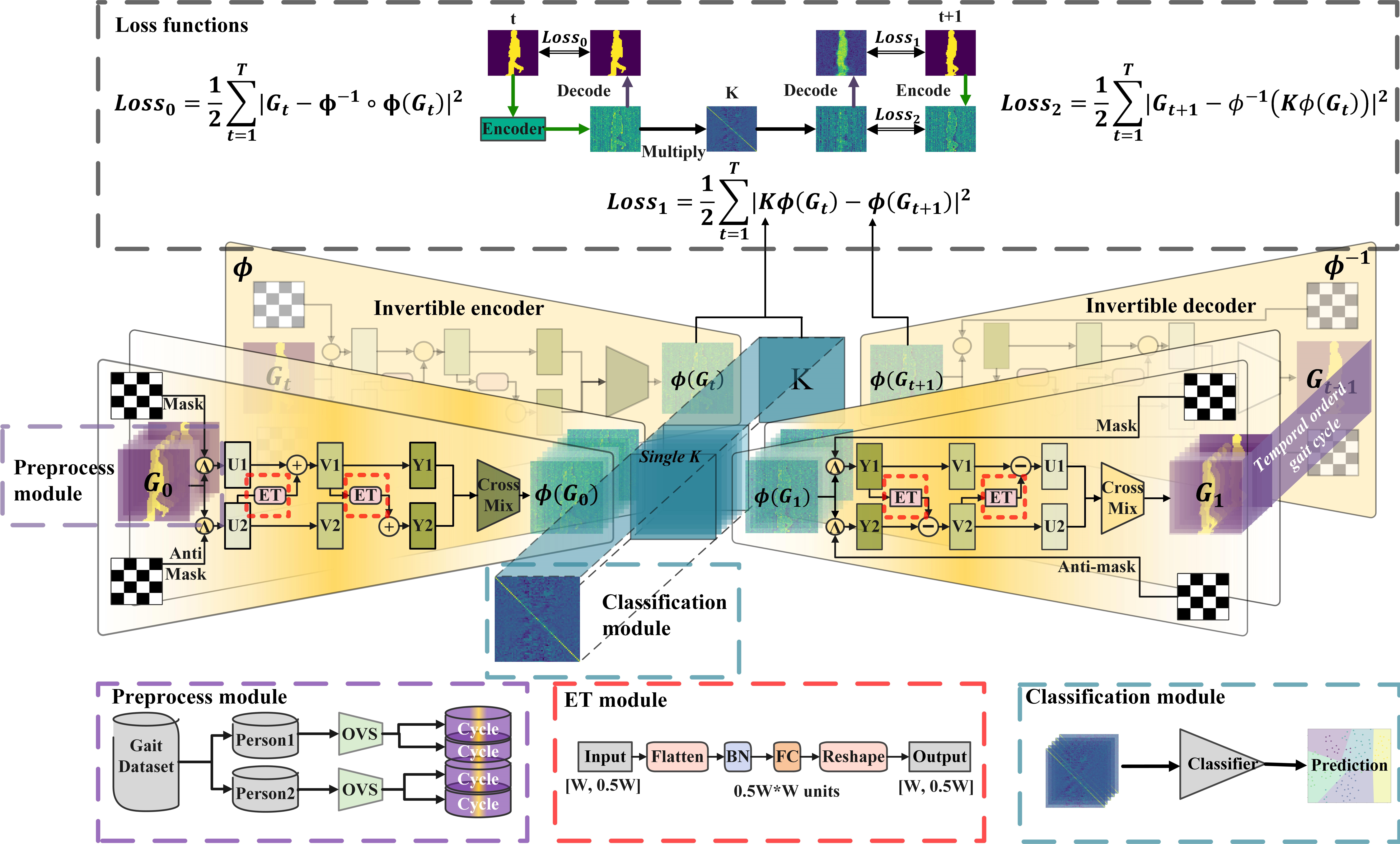}
\caption{the entire structure of our InvKA, where OVS means optimal video segment method, ET means equally transform, BN means batch normalization layer, and DN means full-connect(dense) layer with specific units.}
\label{fig:fullStructue}
\end{figure*}
This chapter provides a detailed explanation of the computational methods used in the InvKA model. Section \ref{period analysis} introduces the OVS method used in the preprocessing process, which periodically segments the gait cycles in the gait dataset (with CASIA-B as the standard). In Section \ref{Koopman analysis}, the gait data is analyzed using the Koopman operator theory to explain the physical significance of this method. Then, in Section \ref{Inertible}, we present the structure and principles of the reversible autoencoder to ensure the completeness, and discuss the difference in computational complexity between using convolutional layers and fully connected layers. Section \ref{training_st} demonstrates the three loss functions used by the model and the specific training strategy.
\par
Fig. \ref{fig:fullStructue} illustrates all the modules of the InvKA model. The execution strategy involves first cutting the gait cycle through the preprocessing module. Then, all video frames are encoded using the same autoencoder, where the ET module includes fully connected and batch normalization layers. The encoded images are used to calculate the Koopman operator $K$ for each gait cycle. Finally, the classification module flattens and classifies $K$ to produce recognition results.
\subsection{Periodic preprocessing}
\label{period analysis}

\par
To ensure that the extracted features are not distorted due to differences in temporal granularity, it is necessary to extract equal-length gait cycles from the original samples in the dataset We have developed a robust algorithm for identifying the gait cycle from the video stream dataset, named OVS(Optimal Video Segment) method. First, we retain only the pose information in the gait stream frames and then reshape them into smaller shapes with a height and width of $w$. Next, we traverse each frame sequentially, calculating the similarity between the benchmark frame and each of the other frames.
\par

In addition, selecting the benchmark frame with the highest variance can clarify the segmentation process for distributions with higher variance. After removing some of the too-close local maximum values, the remaining local maximums are used to determine the best segment positions. In this process, we define the benchmark frame as $M$, and each of the other frames are labeled as $F$. It is important to note that each value in frames $M$ and $F$ is binary.
\par
The similarity $S$ between frame $F$ and the benchmark is:
\begin{equation} \label{Eq:similarity}
	\begin{split}
		\begin{aligned}
			s(F)=\frac{1}{w^2}\sum_{i}^{w}\sum_{j}^{w}[M_{ij}(\textbf{1}-F_{ij})+F_{ij}(\textbf{1}-M_{ij})]
		\end{aligned}
	\end{split}
\end{equation}
\par
\begin{figure}[!htb]
	\centering
	\includegraphics[width=0.45\textwidth]{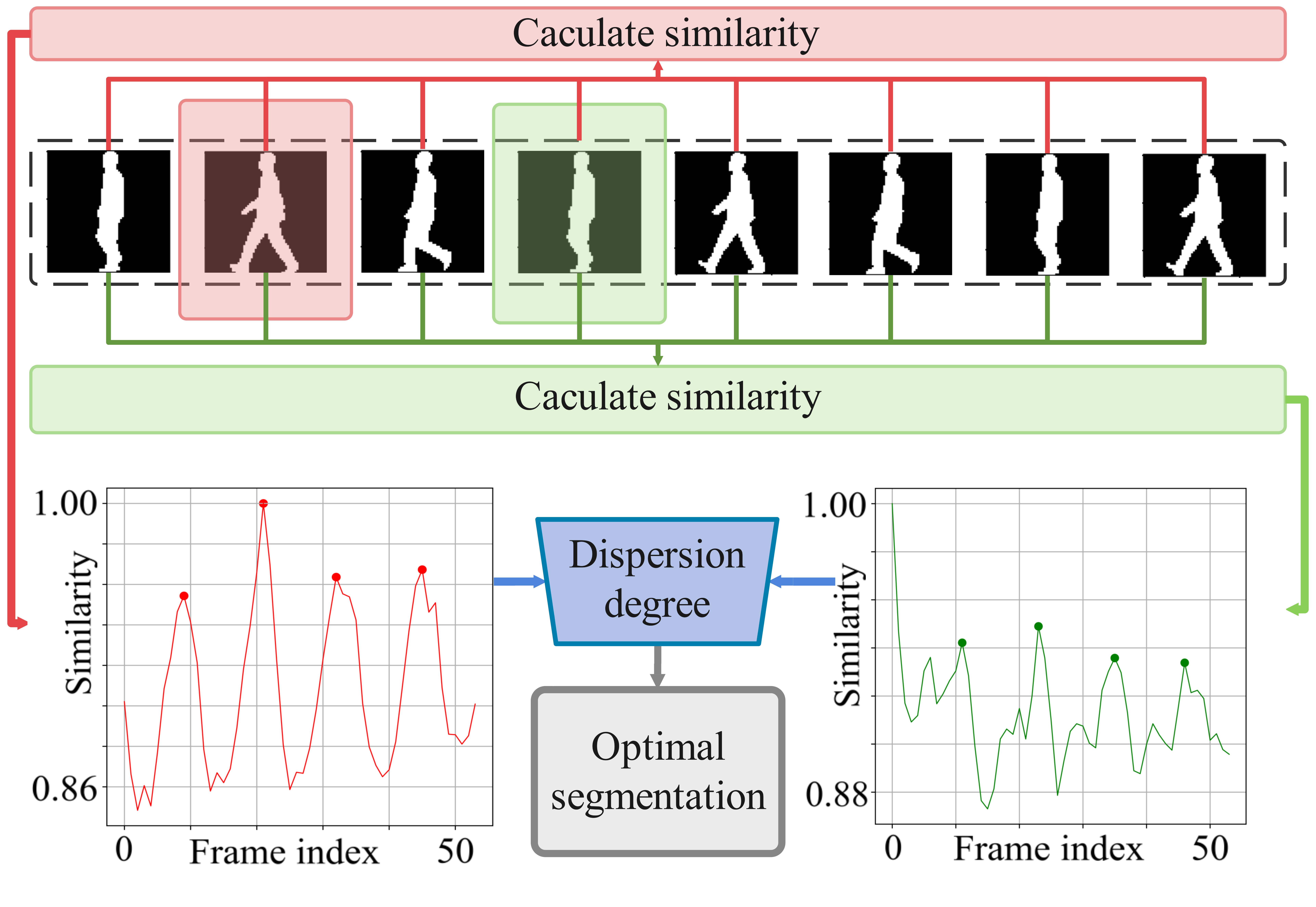}
	\caption{\label{fig:OSS}OVS method, red distribution has higher variance and clearer segment positions.}
\end{figure}

The segment position can be determined by identifying the maximum values. To maintain the consistency of the temporal dimension's sample depth, we remove the maximum values too close to each other. This is because the standard cycle length is set to $T$. Finally, we generate structured data with a size of $[T,w,w]$ as a gait cycle sample. As shown in Fig. \ref{fig:OSS} the left frame is used as the benchmark frame because of its clearer similarity distribution.
\par 
In real-world scenarios, individuals may have different walking speeds, resulting in varying gait cycle lengths. However, the Koopman operator can extract motion information from individuals, and motion information generated by each individual in a similar starting point and of the same cycle length should be symmetric. Therefore, the gait information in each cycle sample is guaranteed to be symmetrical, making them suitable for training.
\subsection{Koopman analysis for frame sequence}
\label{Koopman analysis}
For a gait cycle $\textbf{G}$ in the Spatio-temporal space $ \mathbb{G}:g^{w*w}_{t+1}=\gamma_{t}g^{w*w}_{t} $, where $t(\leq{T})$ represents time, $ w*w $ is the size of bounding box of the gait frames, $ G $ is the gait descriptor, $f_{t}$ is the state transition function from time $t$ to time $t+1$.
\par 
Suppose $\mathbb{L}$ is a Hilbert space,  $\mathbb{L}:X^{w*w}_{t+1}=k_{t}X^{w*w}_{t}$, where $t$ represents time, $ w*w $ is the size of frames, $ X $ is descriptor in Hilbert space, $h_{t}$ is the state transition function from time $t$ to time $t+1$. When $k_{t}$ is an invariant function, the Hilbert space of $\mathbb{L}$ is a linear space. 
\par 
The mapping function\footnote{The $\circ$ operator is an abbreviation for function nesting, $k\circ \phi(x)=k(\phi(x))$} is defined as $\phi$, ${X}$ and ${G}$ are feature matrices in two spaces.
\begin{equation} \label{Eq:phi}
	\begin{split}
		\begin{aligned}
			{\phi}:\mathbb{G} \mapsto  \mathbb{L} \Rightarrow X=\phi(G)
			\Rightarrow \phi(G^{w*w}_{t+1})={k}\circ{\phi}(G^{w*w}_{t})
		\end{aligned}
	\end{split}
\end{equation}

\par 
We refer to the transformed linear space as the embedding space. Typically, the embedding space is an infinite-dimensional linear space. Fortunately, the periodic nature of gait videos suggests that the embedding space for gait videos may be restricted to finite dimensions. In this work, we set the embedding space to the same dimension as the gait video. This enables us to represent the gait video in the embedding space. We denote the matrix format of the function $k$ as $K$.
\par
For each frame $X$ in the embedding space:

\begin{equation} \label{Eq:phiXt}
	\begin{split}
		\begin{aligned}
			\phi(X_{t})=K \phi(X_{t-1})=K^{2} \phi(X_{t-2})=\cdots=K^{t} \phi(X_{0})
		\end{aligned}
	\end{split}
\end{equation}

\par 
Therefore, $K$ contains the kinematic information in the video sequence. Suppose the $K$ of gait video in embedding space has a set of eigenvalues (For non-square matrix, the eigenvalues can be calculated by singular value decomposition) $ {\lambda_{1},\lambda_{2},\lambda_{3},\dots,\lambda_{n}} $, the representation of gait video in embedding space can be simplified to:
\begin{equation} \label{Eq:phiXtlambda}
	\begin{split}
		\begin{aligned}
			\phi(X_{t})=\sum_{i=1}^{n} \lambda_{i}^{t} \phi(X_{0})
		\end{aligned}
	\end{split}
\end{equation}
\par 
These eigenvalues evolve over time with their frequency and decay rate respectively given by $ \angle \lambda_{i}$ and $ \left | \lambda_{i} \right | $. Therefore, the eigenvalues in the embedding space can be investigated to understand the dominant kinematic features of gait videos. 
\subsection{Invertible autoencoder}
\label{Inertible}
\par
Koopman operator theory requires a complete mapping function such that the video data can be mapped frame by frame in the feature space $\mathbb{G}$ and the linear space $\mathbb{L}$:
\begin{equation} \label{Eq:enc}
	\begin{split}
		\begin{aligned}
			\forall \textbf{G}\in \mathbb{G}, \phi^{-1}\circ \phi(\textbf{G})=\textbf{G}
		\end{aligned}
	\end{split}
\end{equation}
\begin{equation} \label{Eq:dec}
	\begin{split}
		\begin{aligned}
			\forall \textbf{X}\in \mathbb{L}, \phi\circ \phi^{-1}(\textbf{X})=\textbf{X}
		\end{aligned}
	\end{split}
\end{equation}
\par
To ensure the validity of formulas (\ref{Eq:enc}) and (\ref{Eq:dec}), the autoencoder must be first trained on the training set $\textbf{G}{tr}$, with formula $Loss_0$ (as described in \ref{training_st}) used to ensure decoding accuracy. This ensures that (\ref{Eq:invert_train}) is satisfied. However, when the encoder and decoder are subsequently frozen and used to perform a spatial transformation on the test set $\textbf{G}{te}$, accuracy may suffer. The decoder, having been trained on $\textbf{G}{tr}$, may not accurately decode features in the embedding space of $\textbf{G}{te}$, leading to an inaccurate feature matrix.

\begin{equation} \label{Eq:invert_train}
	\begin{split}
		\begin{aligned}
			\forall g\in \textbf{G}_{tr}, \phi^{-1}\circ \phi(g)=g
		\end{aligned}
	\end{split}
\end{equation}
\begin{equation} \label{Eq:invert_test}
	\begin{split}
		\begin{aligned}
			\forall g\in \textbf{G}_{te}, \phi^{-1}\circ \phi(g)=g
		\end{aligned}
	\end{split}
\end{equation}
\par
To ensure equation (\ref{Eq:invert_test}), a reversible autoencoder structure is used in the model. This structure allows the model to train only the autoencoder $\phi$ using $\textbf{G}{tr}$, and the exact decoder $\phi^{-1}$ is automatically generated by the autoencoder. This method guarantees accurate coding and decoding in $\textbf{G}{te}$, which ensures that the feature matrix $K$ accurately approximates the kinematic features in the gait video. Therefore, the descriptors in the embedding space can uniquely represent the kinematic features in the gait video. Fig. \ref{fig:fullStructue} shows the autoencoder structure.
\par
The autoencoder splits each frame $G^{w*w}$ in a gait cycle sample into two parts ($U_1$ and $U_2$) using a checkerboard mask. This ensures that both data streams contain enough image feature information to distribute the complexity of information transformation between the functions $f$ and $g$. After the forward conversion, the two frames $ Y_1 $ and $ Y_2 $ are coded from:
\begin{equation} \label{Eq:encoder}
	\begin{split}
		\begin{aligned}
			Y_1&=U_1+f(U_2)\\
			Y_2&=U_2+g(U_1+f(U_2))
		\end{aligned}
	\end{split}
\end{equation}

\par 
Where $f$ and $g$ are arbitrarily complex functions, the function itself does not need to be reversible. However, $f$ and $g$ must have the same shape input and output. We call these functions equally transform(ET).
\par 
The decoding process is the inverse of the encoding process. Frames in the embedding space are first divided into two pieces ($Y_1'$ and $Y_2'$) in the form of a checkerboard used in the coding process. Then the two frames $ U_1' $ and $ U_2' $ are decoded from:

\begin{equation} \label{Eq:decoder}
	\begin{split}
		\begin{aligned}
		    U_2'&=Y_2'-g(Y_1')\\
		    U_1'&=Y_1'-f(Y_2'-g(Y_1'))\\
		\end{aligned}
	\end{split}
\end{equation}

\par 
Therefore, the encoded and decoded image frames are surjective, meaning they can be uniquely represented in embedding space. This property ensures that any data generated by the original autoencoder (regardless of whether it has been used for training before) can be perfectly decoded. During the training process, the encoder and decoder share the layers in the ET functions, resulting in only half the gradient computing cost and parameter count.
InvKA is not an end-to-end model, which means that the algorithm incurs some floating-point operations when slicing gait videos\footnote{These operations can be approximated as $C^{2}_{L}W^2\approx0.01GFLOPs$, where $L$ is the number of frames in the video stream, and $C$ is the combination operator.}.
\par
For the encoding and decoding process, the autoencoder can use a convolution-deconvolution layer structure \cite{zhang2021cross}, or it can choose to flatten the image and process it using fully connected layers. These two methods result in significantly different computational cost. Formula (\ref{Eq:FLOPs_2D}) gives the relationship between the FLOPs and hyperparameters for two types of convolutional layers and fully connected layers. Here, $C_{in}$ is the input channel, $C_{out}$ is the output channel, $W$ and $H$ are the width and height of data, respectively. And $T$ is the time length, $I$ and $O$ are the input and output nodes count, respectively. And $N$ is the kernel size of convolutional layers.
\begin{equation} \label{Eq:FLOPs_2D}
	\begin{split}
		\begin{aligned}
			FLOPs_{ 2D}&=2C_{in}N^2C_{out}WH\\
			FLOPs_{ 3D}&=2C_{in}N^3C_{out}WHT\\
                FLOPs_{ FC}&=2IO
		\end{aligned}
	\end{split}
\end{equation}

In methods using 2D convolution operations, a typical convolutional layer that follows two pooling layers has the following hyperparameters: $C_{in}=64$, $C_{out}=128$, $N=3$, $W=64$, $H=44$. After pooling, the length and width of the image are each reduced by a factor of 4. The ratio of the computational cost between the ET module and this typical convolutional layer is:
\begin{equation} \label{Eq:Comp}
	\begin{split}
		\begin{aligned}
                \frac{FLOPs_{ FC}}{FLOPs_{ 2D}}&=\frac{2IO}{2C_{in}N^2C_{out}\frac{1}{4}W\frac{1}{4}H}\\
                &=\frac{16WH}{C_{in}N^2C_{out}} = 0.611 < 1
		\end{aligned}
	\end{split}
\end{equation}
Thus, the computational cost of the ET module is less than that of a single large-scale convolutional layer. In advanced gait recognition methods, these convolutional layers often exist in groups and are accompanied by other convolutional layers(See TABLE \ref{tab:Computation cost discussion}). However, in the InvKA model, the ET module is used only 2 times. As a result, the InvKA model significantly reduces computational cost by abandoning convlutional layers.
\subsection{Training strategy}
\label{training_st}
In this autoencoder structure, three kinds of loss functions are necessary to ensure our method has caught the kinematic information from those gait cycle samples. And $G_{t+1}$ equals $G_{1}$ ensuring the characteristic of periodicity.
\begin{itemize}
	\item Autoencoder restriction:
	\begin{equation} \label{Eq:loss0}
	\begin{split}
		\begin{aligned}
			Loss_0=\frac{1}{2}\sum_{t=1}^{T}\left\|G_t-\phi^{-1}\circ\phi(G_t)\right\|^2
		\end{aligned}
	\end{split}
    \end{equation}

	This loss can ensure the encoder and decoder are reversible where $\phi$ represents the encoder and $\phi^{-1}$ means the decoder and $G_t$ represents a frame in the gait cycle sample.
	\item Linear restriction:
	\begin{equation} \label{Eq:loss1}
	\begin{split}
		\begin{aligned}
			Loss_1=\frac{1}{2}\sum_{t=1}^{T}\left\|\phi(G_{t+1})-K\phi(G_t)\right\|^2
		\end{aligned}
	\end{split}
    \end{equation}
    Restraining that the gait cycle frames have a linear relationship in the embedding space. Furthermore, the kinematic feature matrix of this space is K.
	\item Prediction restriction:
	\begin{equation} \label{Eq:loss2}
	\begin{split}
		\begin{aligned}
			Loss_2=\frac{1}{2}\sum_{t=1}^{T}\left\|G_{t+1}-\phi^{-1}(K\phi(G_t))\right\|^2
		\end{aligned}
	\end{split}
    \end{equation}
    
	The model should also have the ability to predict future states in the original space.
\end{itemize}
\par 
With the reversible $\phi$ and $\phi^{-1}$, $\textbf{M}$ can be any tensor: 

	\begin{equation} \label{Eq:revi}
	\begin{split}
		\begin{aligned}
			\textbf{M}=\phi^{-1}\circ{}\phi(\textbf{M})
		\end{aligned}
	\end{split}
    \end{equation}
When $\textbf{M}$ was set to a frame matrix $G$, with (\ref{Eq:revi}):

	\begin{equation} \label{Eq:loss0_0}
	\begin{split}
		\begin{aligned}
			G=\phi^{-1}\circ{}\phi(G) &\Rightarrow \left\|G-\phi^{-1}\circ{}\phi(G)\right\|^2=0 \\
   &\Rightarrow Loss_0=0
		\end{aligned}
	\end{split}
    \end{equation}
    
When $Loss_1=0$, with (\ref{Eq:revi}) and (\ref{Eq:loss0_0}):

	\begin{equation} \label{Eq:loss1=0As}
	\begin{split}
		\begin{aligned}
			\phi(&G_{t+1})=K\phi(G_t) \Rightarrow \phi^{-1}\circ{}\phi(G_{t+1}) \\&= \phi^{-1}(K\phi(G_{t})) \Rightarrow G_{t+1}-\phi^{-1}(K\phi(G_t))=0 \\&\Rightarrow Loss_2=0
		\end{aligned}
	\end{split}
    \end{equation}
Thus, if $\phi$ and $\phi^{-1}$ are reversible, $Loss_1=0$ is a sufficient condition for $Loss_2=0$. Therefore, according to the reversible neural network architecture, using only $Loss_1$ is equivalent to three loss functions, detailed ablation experiments are presented in Section \ref{ablation}.
This model consists of two training modules: the encoder/decoder pair $\phi$ and $\phi^{-1}$, and the kinematic feature matrix $K$. Based on Koopman analysis, \textbf{$\phi$ and $\phi^{-1}$ are shared by all samples}, ensuring that all samples are treated equally during the training process. After training, an autoencoder is generated to convert video stream space to embedding space for the entire gait set. Additionally, the coder-training process also trains the kinematic feature matrix $K$. This pre-trained $K$ can serve as a \textbf{prototype} to shorten the matrix-training process.
\par
\begin{figure}[!h]
\centering
\includegraphics[width=0.45\textwidth]{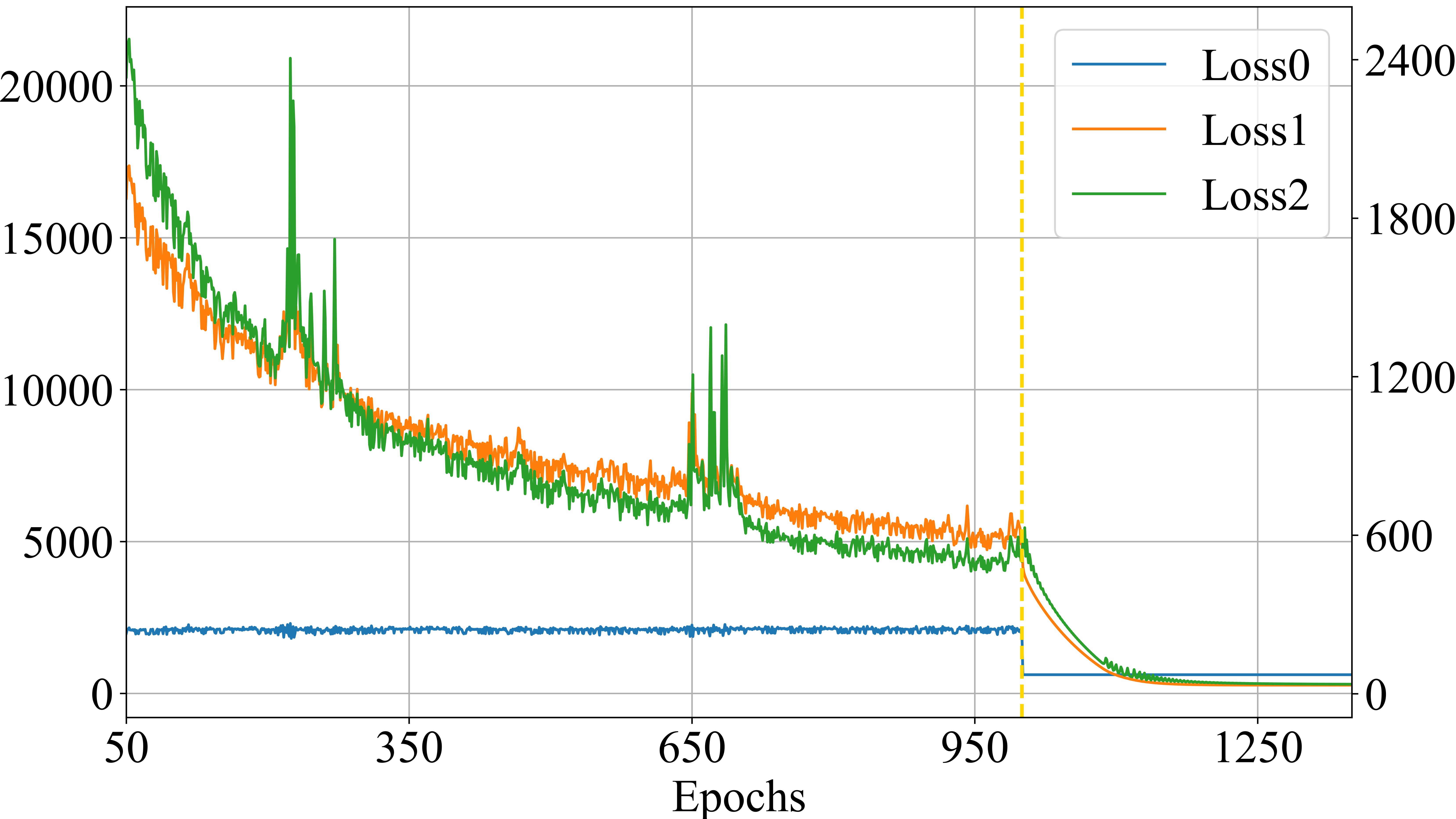}
\caption{\label{fig:Training_losses}training losses, left to the yellow vertical line is the coder-training process, and on the right side is the matrix-training process.}
\end{figure}

Training of $K$ adopts a gait-cycle oriented specificity training approach, where each gait cycle produces an individual kinematic feature matrix in the embedding space. All layers except $K$ are frozen, and only $Loss_1$($L$) is used to compute the gradient, making this an optimization problem for $K$ to $L$. The gait cycle tensor with shape $[T,W,W]$ is represented by $\textbf{G}$.
In Appendix C, it can be proven that:
\begin{equation} \label{Eq:secpartial3}
	\begin{split}
		\begin{aligned}
		    \frac{\partial^2 L(\textbf{G},\phi, K)}{\partial K^2} \geq 0\\
		\end{aligned}
	\end{split}
\end{equation}
Thus, the loss function $L$ is a convex function with respect to $K$, making the optimization problem of training $Loss_1$ a smooth and stable gradient descent process. This also ensures that the obtained $K$ value is globally optimal, making the training process highly reliable. Fig. {\ref{fig:Training_losses}} displays the training losses throughout the entire process. After training, a dataset containing a set of kinematic feature matrices with shape $[w,w]$ is generated, where each matrix corresponds to a gait cycle and a single person. 
\par
Moreover, we provide a \textbf{closed-form solution} for the minimum least-squares solution of the convex optimization problem, which is the \textbf{analytical solution} for $K$ when $L$ is minimized(See Appendix D). Using the closed-form solution eliminates the need to calculate each kinematic feature matrix $K$ individually, thus enabling the possibility of batch operations for this approach. However, to ensure the experimental rigor, we still use the training strategy of freezing certain layers to discuss the model's performance in subsequent processes.

\section{Performance}
In this chapter, the InvKA model is evaluated from the perspectives of interpretability and computational complexity. The CASIA-B dataset is introduced in Section \ref{dataset}. In Section \ref{visual}, the interpretability of the InvKA model is evaluated using the PDR framework. In Section \ref{computional cost}, the model’s complexity and accuracy are compared to optimize computational cost. In Section \ref{validation}, the recognition accuracy of the model on the non-occlusion dataset (CASIA-B NM) is compared with state-of-the-art models. 
In Section \ref{ablation}, the choice of specific components in the model is explained through ablation experiments. Section \ref{Implementation} describes the hyperparameter configuration of the InvKA model. 
\subsection{Dataset introduction}
\label{dataset}
The CASIA-B dataset is widely used in gait recognition research, with about 80\% of researchers choosing it to evaluate their methods\cite{9714177}. In this study, this dataset is used to evaluate the effectiveness of our approach. The dataset includes walking posture image data for 124 individuals and is divided into three categories: Normal (NM), Bag (BG), and Clothes (Cl). The NM category provides 6 video frame sequence samples per person, while the BG and Cl categories each provide 2 video samples.
\subsection{Interpretability evaluation}
\label{interpretability}
\par
WJ Murdoch et al. come up with a framework to evaluate model's interpretability\cite{murdoch2019definitions}, and that PDR(Predictive, Descriptive,Relevant) framework is used to evaluate our model's interpretability.
\par Five views of the proposed model are discussed:
\par
\begin{figure}[!htb]\begin{center}
        \includegraphics[width=0.45\textwidth]{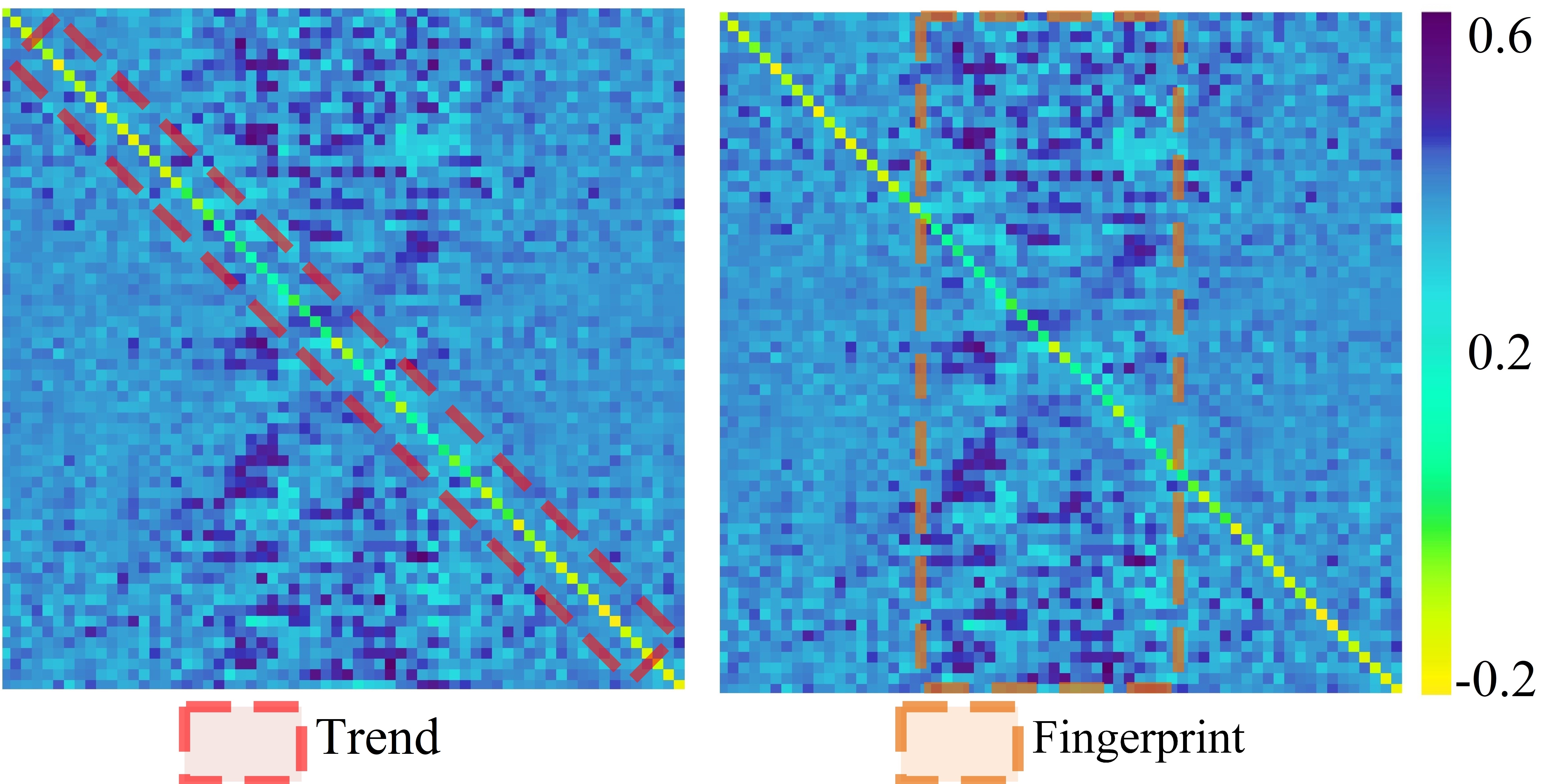}
        \caption{\label{fig:K}the two figures show the kinematic feature matrix extracted from the same gait cycle. The gait trend information is contained in the diagonal elements, while the gait fingerprint data is reflected in the middle part.}
\end{center}\end{figure}
\begin{figure}[!htb]
\centering
\includegraphics[width=0.45\textwidth]{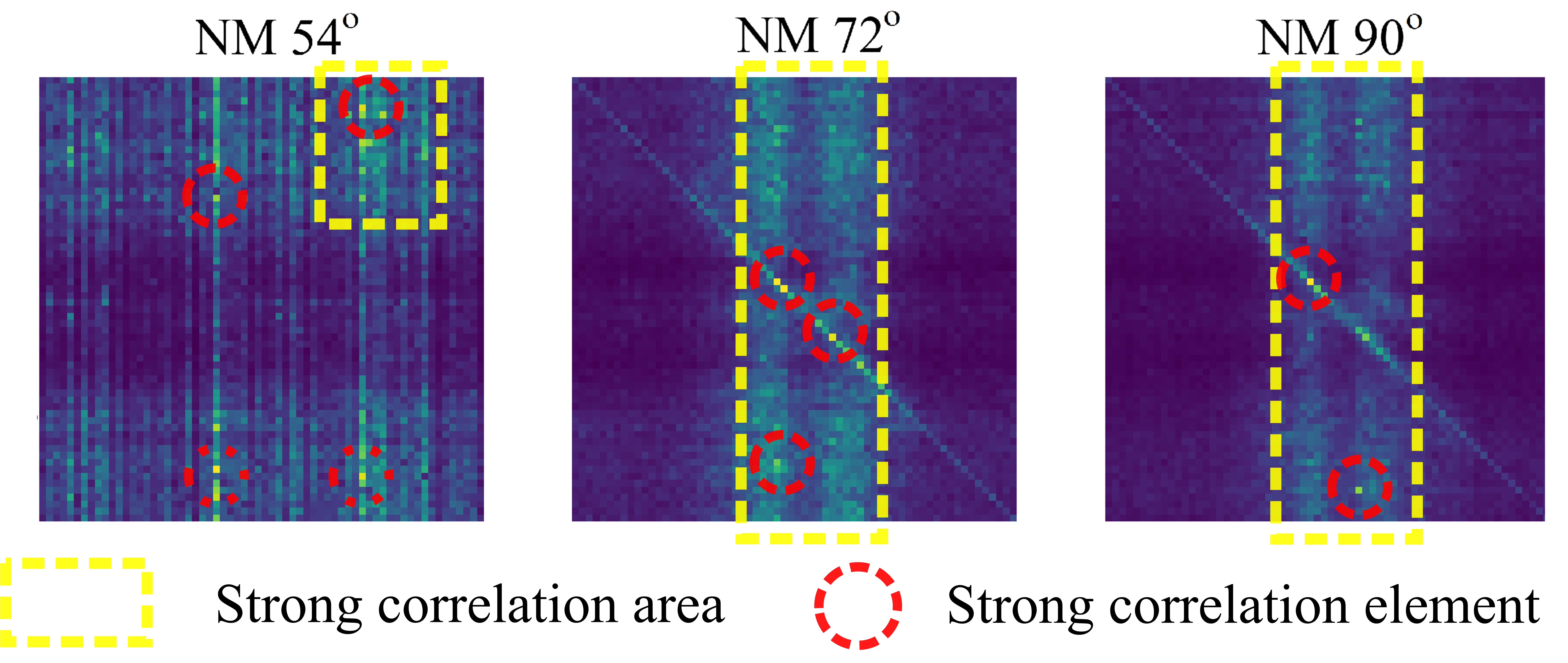}
\caption{\label{fig:Coef_type}the weight matrix of a logistic regression classifier under different angle datasets. As the angle approaches $90^\circ$, the strong correlation elements are more likely to be distributed on the diagonal elements.}
\end{figure}

\textbf{\emph{Visualization}}
\label{visual}
Fig. \ref{fig:K} shows the information contained in the kinematic feature matrix. The diagonal elements in the heatmap contain prominent trend information in posture changes, such as large arm swings and leg movements. The elements in the middle of the heatmap on the right represent the trend information of gait features. These details may not be directly discernible to the human eye, just like fingerprint. However, these fingerprint feature play a dominant role in the classification results. A detailed discussion of the focus of the classifier is presented in Fig. \ref{fig:Coef_type}.
\par
\textbf{\emph{Modularity}}: The modular model is designed to have a meaningful prediction process with the independent interpretation of each step, which facilitates reasoning about the different parts of the model. As illustrated in Fig. \ref{fig:fullStructue}, the model adopts a modular structure, where different algorithms are used for segmentation, coding and decoding, and classification tasks. Segmentation divides a large temporal video sequence into small gait cycles with a fixed length. The autoencoder outputs the kinematic feature matrix, and finally, the classifier performs the classification. A comparison of different algorithms for the classifier has been discussed in \ref{ablation}.

\par
\begin{figure}[htb]
    \begin{center}
        \includegraphics[width=0.4\textwidth]{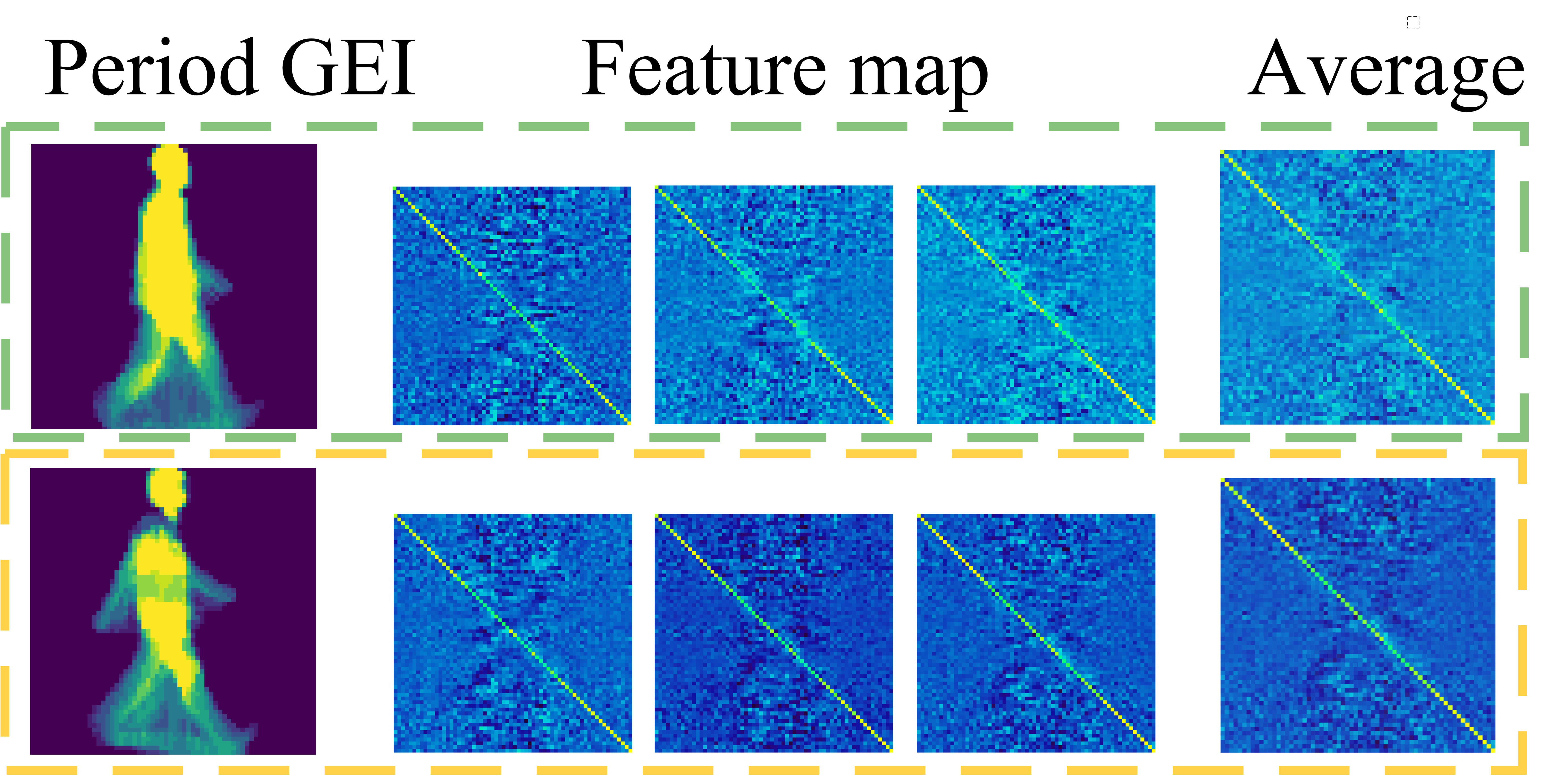}
        \caption{\label{fig:K-NM}kinematic feature matrix for NM $90^\circ$ dataset from two persons.}
    \end{center}
\end{figure}

\textbf{\emph{Simulatability}}: Simulability refers to the ability of the interpretable model to internally simulate and reason about its decision-making process. However, as the complexity of the model increases, it becomes increasingly difficult to simulate the internal workings of the model. As shown in Fig. \ref{fig:performance}, our model has a significant advantage in terms of complexity compared to other models. Despite this, the number of floating-point operations performed by the model remains relatively high, indicating that the model is still far from being fully simulable.
\textbf{\emph{Model-Based feature engineering}}: The heat map shown in Fig. \ref{fig:K-NM} displays the feature matrix of six gait cycles from two individuals. While it may be difficult to differentiate the detailed features of the six samples with the naked eye, the feature matrix is flattened into a vector and input into a logistic regression classifier, yielding highly accurate classification results similar to trend identification. It is important to note that each of the feature matrix heat maps is physically meaningful.
\par
\label{Prediction-interpretation}
\textbf{\emph{Prediction-level interpretation}}: A logistic regression classifier is employed to classify the flattened matrix $K$, which can be considered as a neural network with a single hidden layer. To deal with the vast number of features, we set a minimal regularization weight to ensure that the classifier had adequate complexity to fit our classification model. Following classification, we obtained an array $C$, where each value corresponds to the weight of one element for a given class\footnote{$\beta$ refers to the bias, and $F()$ is the flatten operation.}.
\begin{equation} \label{Eq:second code INN}
	\begin{split}
		\begin{aligned}
			y_{pred} = \sigma(C^TF(K)\pm\beta)
		\end{aligned}
	\end{split}
    \end{equation}
\par
The sigmoid function denoted by $\sigma$ is monotonic and the bias $\beta$ is independent of input parameters. During the classification process, each element in the vector $C$ is multiplied by each element in the spreading $K$. After the fitting process, the vector $C$ is reshaped into a matrix with shape $[W, W]$, where the absolute value of each element corresponds to the weight of a element in the Koopman operator. A higher weight indicates a strong correlation between that element and the label.
\par
Based on the weight matrix shown in Fig. \ref{fig:Coef_type}, higher values are mainly located on the diagonal of the kinematic feature matrix when the shooting angle is close to $90^\circ$. This indicates that the classifier focuses more on the diagonal elements, which are closely related to the classification categories. However, the diagonal elements are not decisive in the classification process, as there are a large number of off-diagonal elements that have a significant impact on the classification, despite their smaller quantity. This suggests that the off-diagonal elements of the matrix still contain important gait information that influences the classification process. The weight matrices for all galleries are shown in Appendix E. 
\subsection{Computational cost}
\label{Computational cost}
\begin{figure}[htb]
\centering
\includegraphics[width=0.35\textwidth]{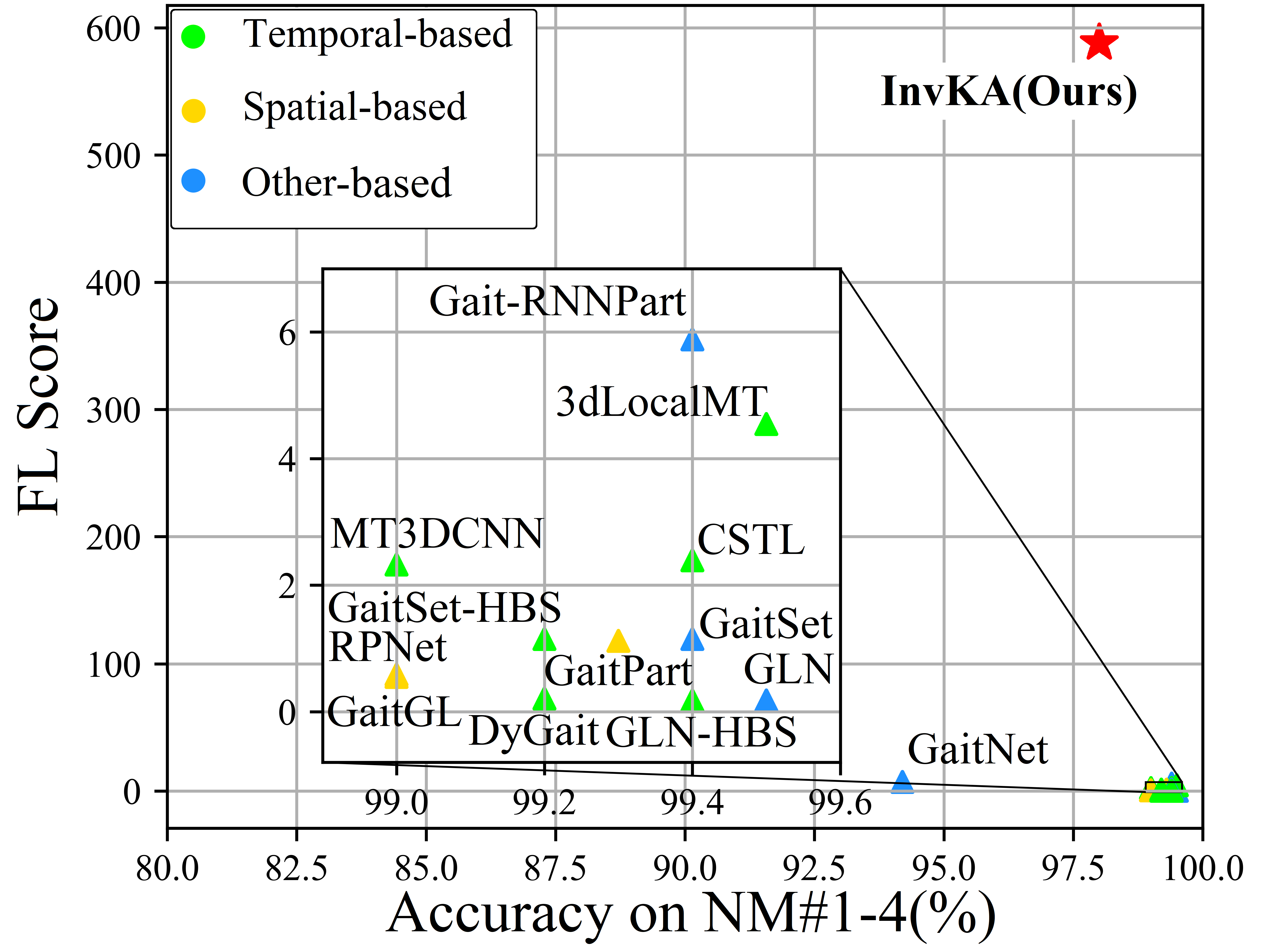}
\caption{\label{fig:Cost-acc}compared to recent models, the complexity of InvKA is at least two orders of magnitude smaller than that of recent models.}
\end{figure}

\begin{table*}[!htb]
\centering
\caption{performence discussion on state-of-the-art methods}
\label{tab:Computation cost discussion}
\setlength{\tabcolsep}{0.5mm}{
\begin{tabular}{c|c|c|cc|cc|c}
\hline
Methods      & Year & Venue       & \multicolumn{2}{c|}{Structure}         & \multicolumn{2}{c|}{Performence}                                                                         & Method  \\ \hline
Compoment    &      &             & \multicolumn{1}{c|}{\begin{tabular}[c]{@{}c@{}}2D-Conv \\ ($\ge$)\end{tabular}} & {\begin{tabular}[c]{@{}c@{}}3D-Conv \\ ($\ge$)\end{tabular}} & \multicolumn{1}{c|}{\begin{tabular}[c]{@{}c@{}}Complexity \\ (GFLOPs)\end{tabular}} & \begin{tabular}[c]{@{}c@{}}Accuracy \\ (LT \%)\end{tabular} &                     \\ \hline
GEI          & 2008 & -           & \multicolumn{1}{c|}{2}       & 0       & \multicolumn{1}{c|}{13}                                                               & -                & Energy map                  \\
Gait-RNNPart  & 2020 & T-BIOM      & \multicolumn{1}{c|}{6}       & 0       & \multicolumn{1}{c|}{1.7}                                                              & 99.4             & Recurrent learning          \\
MT3DCNN       & 2020 & ACM-MM      & \multicolumn{1}{c|}{0}       & 12      & \multicolumn{1}{c|}{4.3}                                                              & 99.0             & Temporal-based               \\
GLN           & 2020 & ECCV        & \multicolumn{1}{c|}{16}      & 0       & \multicolumn{1}{c|}{56}                                                               & \textbf{99.5}            & Pyramid-mapping              \\
GaitPart     & 2020 & CVPR        & \multicolumn{1}{c|}{26}      & 0       & \multicolumn{1}{c|}{8.93}                                                             & 99.3             & Spatial-based               \\
GaitGL        & 2021 & ICCV        & \multicolumn{1}{c|}{0}       & 5       & \multicolumn{1}{c|}{18.2}                                                             & 99.0             & Spatial-based               \\
3dLocalMT     & 2021 & ICCV        & \multicolumn{1}{c|}{0}       & 6       & \multicolumn{1}{c|}{2.2}                                                              & 99.5             & Temporal-based             \\
CSTL          & 2021 & ICCV        & \multicolumn{1}{c|}{4}       & 2       & \multicolumn{1}{c|}{4.18}                                                             & 99.4            & Temporal-based               \\
OFA          & 2021 & CVPR        & \multicolumn{1}{c|}{6}       & 0       & \multicolumn{1}{c|}{1.7}                                                              & -                & Koopman-based                     \\
RPNet         & 2021 & IEEE CSVT   & \multicolumn{1}{c|}{21}      & 0       & \multicolumn{1}{c|}{17}                                                               & 99.1             & Spatial-based                \\
GaitSet      & 2022 & IEEE T-PAMI & \multicolumn{1}{c|}{10}     & 0       & \multicolumn{1}{c|}{8.7}                                                              & 99.4            & Pyramid-mapping              \\
GaitNet       & 2022 & IEEE T-PAMI & \multicolumn{1}{c|}{8}       & 0       & \multicolumn{1}{c|}{1.4}                                                              & 95.1             & Auto-encoder                     \\
DyGait        & 2023 & Arxiv       & \multicolumn{1}{c|}{0}       & 10      & \multicolumn{1}{c|}{48}                                                               & 99.2            & Temporal-based               \\
GLN-HBS       & 2023 & WACV        & \multicolumn{1}{c|}{16}     & 0       & \multicolumn{1}{c|}{56}                                                               & 99.4             & Temporal-based             \\
GaitSet-HBS   & 2023 & WACV        & \multicolumn{1}{c|}{10}     & 0       & \multicolumn{1}{c|}{8.7}                                                              & 99.2             & Temporal-based               \\
\textbf{InvKA(Ours)}         &      &             & \multicolumn{1}{c|}{$\textbf{0}$ }       & $\textbf{0}$       & \multicolumn{1}{c|}{$\textbf{0.017}$}                                                            & \textbf{98.0}             & \textbf{Koopman-based}                    \\ \hline
\end{tabular}}
\end{table*}

\label{computional cost}

\par
Due to the significantly lower computational complexity of fully connected layers compared to convolutional layers, we estimate the FLOPs of the models solely based on convolutional layers. Additionally, we also ignore preprocessing and detailed operations if their magnitudes are too small. As the difference in FLOPs between models are often on the order of magnitude, we have defined a score, FL Score, to intuitively demonstrate the gap in floating-point operations:
\begin{equation} \label{Eq:FL Score}
	\begin{split}
		\begin{aligned}
			FLScore = \frac{10^{8}}{FLOPs}
		\end{aligned}
	\end{split}
\end{equation}
\par
As shown in Fig. \ref{fig:Cost-acc}, our model has reduced the FLOPs by more than 100 times compared to other models, including the SOTA model. Meanwhile, the optimal angle accuracy of our model still achieved a competitive level (98\%).
\begin{figure}[htb]
\centering
\includegraphics[width=0.4\textwidth]{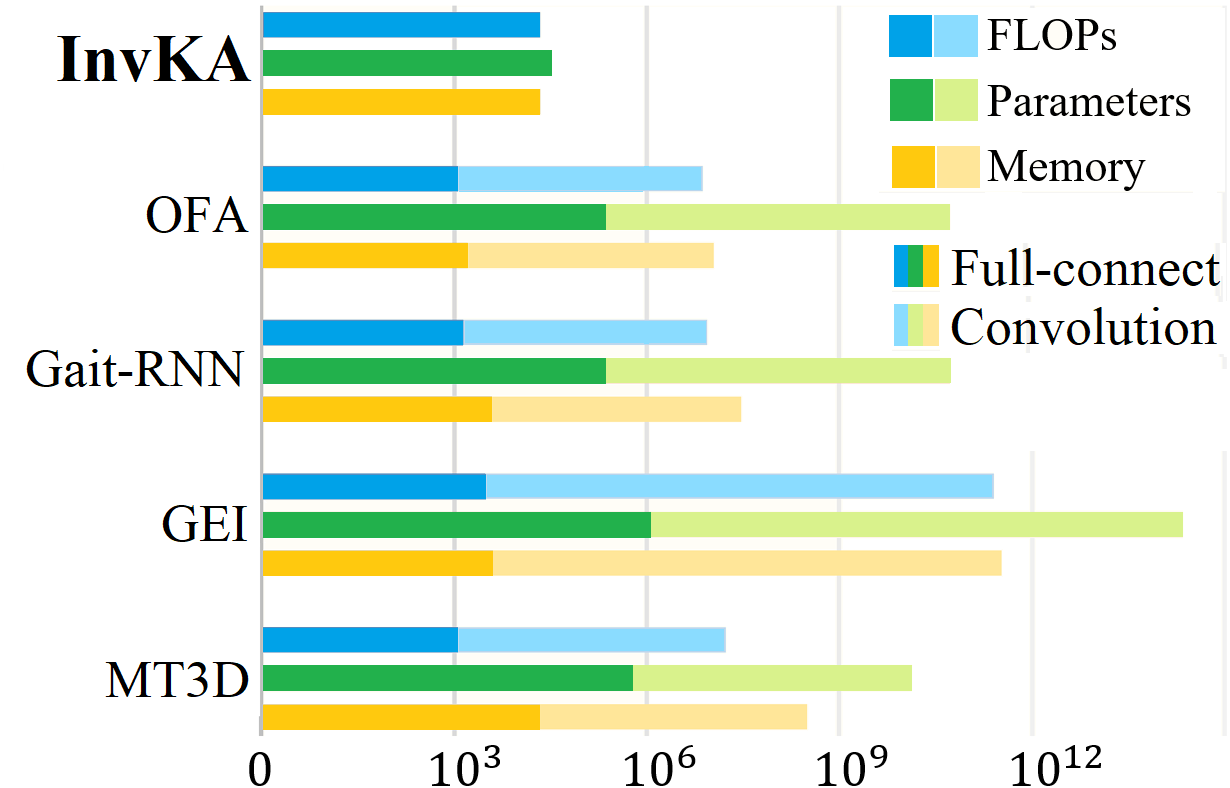}
\caption{\label{fig:performance}computational cost comparison, the dark part shows the computational cost of the full-connect layer, and the light part shows the computational cost of the convolutional layer. And the unit of column height in the figure is 1.}
\end{figure}
Fig.  \ref{fig:performance} shows the performance evaluation of model components, the dark part shows the computational cost of the full-connect layer, and the light part shows the computational cost of the convolutional layer. GEINet \cite{yoo2008automated} represents the gait energy figure method, Gait-RNNPart \cite{sepas2020view} represents the 2D-convolution method, and MT3DCNN \cite{lin2020gait} represents the 3D-convolution method, and OFA \cite{zhang2021cross} represents the auto-coding method, which is similar to ours. As can be seen from the figure, the convolutional layers occupy the majority of the computational cost in these models.
\subsection{Validation accuracy}
\label{validation}
TABLE \ref{tab:validation accuracy} presents the performance of our model and other models on non-occlusion dataset of different sizes(ST,MT,LT)\footnote{The abbreviations "S" (Short), "M" (Middle), and "L" (Large) refer to the test protocols used in the experiments, which contain 24, 62, and 74 sample data respectively.}. We evaluate our model's accuracy through small-sample, medium-sample, and large-sample training. Compared with the fairly high-precision models, our model demonstrates higher performance in specific experiments. Additionally, our model does not require extensive data to train deep-layer parameters. We reduce the kinematic detail information in the training data for data from different angles. Moreover, shooting non-opposite angles ($18^\circ,36^\circ,36^\circ,54^\circ,72^\circ,108^\circ$) may inevitably produce redundant information due to the pose size changing, leading to insufficient representation of the kinematic information used for sequence splitting and kinematic feature matrix training.
\begin{table*}[ht]\centering
\caption{recognition rank-1 accuracy (\%) cross views under NM on CASIA-B Dataset($\%$)}
\label{tab:validation accuracy}
\centering
\setlength{\tabcolsep}{3mm}{

\begin{tabular}{ccc|ccccccc}
\hline
\multicolumn{3}{c|}{Gallery NM$\#$1-4}                                                     & \multicolumn{7}{c}{$0^\circ$-$108^\circ$}                                                                                                                                                                                            \\ \hline
\multicolumn{3}{c|}{Probe View}                                                            & $0^\circ$                      & $18^\circ$                     & $36^\circ$                     & $54^\circ$                     & $72^\circ$                     & $90^\circ$                     & $108^\circ$                    \\ \hline
\multicolumn{1}{c|}{}       & \multicolumn{1}{c|}{Method}                           & Year &                                &                                &                                &                                &                                &                                &                                \\ \hline
\multicolumn{1}{c|}{ST(24)} & \multicolumn{1}{c|}{CMCC \cite{kusakunniran2013recognizing}} & 2013 & 46.3                           & -                              & -                              & 52.4                           & -                              & 48.3                           & -                              \\
\multicolumn{1}{c|}{}       & \multicolumn{1}{c|}{CNN-LB \cite{wu2016comprehensive}}       & 2016 & 54.8                           & -                              & -                              & 77.8                           & -                              & 64.9                           & -                              \\
\multicolumn{1}{c|}{}       & \multicolumn{1}{c|}{MT3DCNN \cite{lin2020gait}}              & 2020 & 71.9                           & 83.9                           & 90.9                           & 90.1                           & 81.1                           & 75.6                           & 82.1                           \\
\multicolumn{1}{c|}{}       & \multicolumn{1}{c|}{GaitGL \cite{lin2021gait}}               & 2021 & 77.0                           & 87.8                           & \textbf{93.9} & 92.7                           & \textbf{83.9} & 78.7                           & 84.7                           \\
\multicolumn{1}{c|}{}       & \multicolumn{1}{c|}{GaitSet \cite{9351667}}          & 2022 & 64.6                           & 83.3                           & 90.4                           & 86.5                           & 80.2                           & 75.5                           & 80.3                           \\
\multicolumn{1}{c|}{}       & \multicolumn{1}{c|}{\textbf{InvKA(Ours)}}                      &      & \textbf{94.1} & \textbf{90.9} & \textbf{89.7} & \textbf{93.2} & \textbf{85.1} & \textbf{92.3} & \textbf{93.5} \\ \hline
\multicolumn{1}{c|}{MT(62)} & \multicolumn{1}{c|}{AE \cite{yu2017invariant}}               & 2017 & 49.3                           & 61.5                           & 64.4                           & 63.6                           & 63.7                           & 58.1                           & 59.9                           \\
\multicolumn{1}{c|}{}       & \multicolumn{1}{c|}{MGAN \cite{he2018multi}}                 & 2018 & 54.9                           & 65.9                           & 72.1                           & 74.8                           & 71.1                           & 65.7                           & 70.0                           \\
\multicolumn{1}{c|}{}       & \multicolumn{1}{c|}{MT3DCNN \cite{lin2020gait}}             & 2020 & 91.9                           & 96.4                           & 98.5                           & 95.7                           & 93.8                           & 90.8                           & 93.9                           \\
\multicolumn{1}{c|}{}       & \multicolumn{1}{c|}{GaitGL \cite{lin2021gait}}               & 2021 & 93.9                           & \textbf{97.6} & \textbf{98.8} & \textbf{97.3} & \textbf{95.2} & \textbf{92.7} & \textbf{95.6} \\
\multicolumn{1}{c|}{}       & \multicolumn{1}{c|}{GaitSet \cite{9351667}}          & 2022 & 86.8                           & 95.2                           & 98.0                           & 94.5                           & 91.5                           & 89.1                           & 91.1                           \\
\multicolumn{1}{c|}{}       & \multicolumn{1}{c|}{\textbf{InvKA(Ours)}}                      &      & \textbf{95.0} & 92.3                           & 90.3                           & \textbf{93.1} & 88.0                           & \textbf{97.3} & \textbf{92.9} \\ \hline
\multicolumn{1}{c|}{LT(74)} & \multicolumn{1}{c|}{CNN-3D \cite{wu2016comprehensive}}       & 2016 & 87.1                           & 93.2                           & 97.0                           & 94.6                           & 90.2                           & 88.3                           & 91.1                           \\
\multicolumn{1}{c|}{}       & \multicolumn{1}{c|}{CNN-Ensemble \cite{wu2016comprehensive}} & 2016 & 88.7                           & 95.1                           & 98.2                           & 96.4                           & 94.1                           & 91.5                           & 93.9                           \\
\multicolumn{1}{c|}{}       & \multicolumn{1}{c|}{Gait-RNNPart \cite{sepas2020view}}       & 2020 & 91.1                           & 98.0                           & 99.4                           & 98.2                           & 93.2                           & 91.9                           & 95.2                           \\
\multicolumn{1}{c|}{}       & \multicolumn{1}{c|}{MT3DCNN \cite{lin2020gait}}              & 2020 & 95.7                           & 98.2                           & 99.0                           & 97.5                           & 97.5                           & 93.9                           & 96.1                           \\
\multicolumn{1}{c|}{}       & \multicolumn{1}{c|}{GLN \cite{hou2020gait}}                  & 2020 & 93.2                           & 99.3                           & 99.5                           & 98.7                           & 96.1                           & 95.6                           & 97.2                           \\
\multicolumn{1}{c|}{}       & \multicolumn{1}{c|}{GaitPart \cite{fan2020gaitpart}}         & 2020 & 94.1                           & 98.6                           & 99.3                           & 98.5                           & 94.0                           & 92.3                           & 95.9                           \\
\multicolumn{1}{c|}{}       & \multicolumn{1}{c|}{GaitGL \cite{lin2021gait}}               & 2021 & 96.0                           & 98.3                           & 99.0                           & 97.9                           & 96.9                           & 95.4                           & 97.0                           \\
\multicolumn{1}{c|}{}       & \multicolumn{1}{c|}{3dLocalMT \cite{huang20213d}}            & 2021 & 96.0                           & 99.0                           & 99.5                           & \textbf{98.9} & 97.1                           & 94.2                           & 96.3                           \\
\multicolumn{1}{c|}{}       & \multicolumn{1}{c|}{CSTL \cite{huang2021context}}            & 2021 & 97.8                           & \textbf{99.4} & 99.2                           & 98.4                           & 97.3                           & 95.2                           & 96.7                           \\
\multicolumn{1}{c|}{}       & \multicolumn{1}{c|}{RPNet \cite{qin2021rpnet}}               & 2021 & 95.1                           & 99.0                           & 99.1                           & 98.3                           & 95.7                           & 93.6                           & 95.9                           \\
\multicolumn{1}{c|}{}       & \multicolumn{1}{c|}{GaitSet \cite{9351667}}          & 2022 & 90.8                           & 97.9                           & \textbf{99.4} & 96.9                           & 93.6                           & 91.7                           & 95.0                           \\
\multicolumn{1}{c|}{}       & \multicolumn{1}{c|}{GaitNet \cite{9154576}}                  & 2022 & 93.1                           & 92.6                           & 90.8                           & 92.4                           & 87.6                           & 95.1                           & 94.2                           \\
\multicolumn{1}{c|}{}       & \multicolumn{1}{c|}{DyGait \cite{wang2023dygait}}            & 2023 & \textbf{97.4} & 98.9                           & 99.2                           & 98.3                           & \textbf{97.7} & \textbf{96.8} & \textbf{98.2} \\
\multicolumn{1}{c|}{}       & \multicolumn{1}{c|}{GLN-HBS \cite{zhu2023gait} }             & 2023 & 93.8                           & 98.9                           & 99.4                           & \textbf{98.9} & 95.2                           & 94.2                           & 95.4                           \\
\multicolumn{1}{c|}{}       & \multicolumn{1}{c|}{GaitSet-HBS \cite{zhu2023gait}}          & 2023 & 96.0                           & 98.3                           & 99.2                           & 97.8                           & 96.4                           & 95.9                           & 97.4                           \\
\multicolumn{1}{c|}{}       & \multicolumn{1}{c|}{\textbf{InvKA(Ours)}}                      &      & \textbf{94.3} & 92.4                           & 90.6                           & 91.4                           & 87.4                           & \textbf{98.0} & 91.6                           \\ \hline
\end{tabular}}
\end{table*}
Our model exhibits a noticeable difference from other models in that its performance is significantly better on small datasets than on large datasets, which may seem anomalous. This phenomenon occurs because our model only employs the neural network as a tool to fit the embedding space, while the logistic regression classifier carries out the final classification task. For statistical learning classifiers with low complexity and strong interpretability, larger and more complex sample sizes can easily exceed the classifier's capacity, leading to underfitting.

\subsection{Ablation experiment}
\label{ablation}
The neural network was only used as a tool to fit the intermediate value $K$, not as the final classification method. Therefore, the loss function values are used to evaluate the fitting ability of the neural network. In a fixed-length epoch, the best fitting result $Loss_{all}$(Sum of loss values) is used as the evaluation basis of the model. Oppositely, when evaluating the classifiers, validation accuracy can be used.
\par
\begin{table*}[!htb]
\centering
\caption{ablation experiment}
\begin{small}
\label{tab:Component discussion}
\centering
\setlength{\tabcolsep}{1mm}{
\begin{tabular}{c|ccc|cccccccc|ccc}
\hline
Discussion    & \multicolumn{3}{c|}{ET}                                       & \multicolumn{3}{c|}{Loss}                                                            & \multicolumn{5}{c|}{Classifier}                                                                                    & \multicolumn{3}{c}{Evaluation index}                                       \\ \hline
Components    & \multicolumn{1}{c|}{DN} & \multicolumn{1}{c|}{BN} & Conv2D    & \multicolumn{1}{c|}{$Loss_0$} & \multicolumn{1}{c|}{$Loss_1$} & \multicolumn{1}{c|}{$Loss_2$} & \multicolumn{1}{c|}{DT} & \multicolumn{1}{c|}{SVM} & \multicolumn{1}{c|}{NB} & \multicolumn{1}{c|}{KM} & LR        & \multicolumn{1}{c|}{$Loss_{all}$} & \multicolumn{1}{c|}{Accuracy} & Time(s)     \\ \hline
\textbf{Best} & \checkmark               & \checkmark               &           &                            & \checkmark                  &                            &                         &                          &                         &                         & \checkmark & $1384\pm50$                    & \textbf{96.0$\pm$3.6}           & 103         \\ \hline
ET            & \checkmark               &                         & \checkmark &                            & \checkmark                  &                            &                         &                          &                         &                         & \checkmark & $1710\pm50$                    & -                             & \textbf{92} \\
              & \checkmark               &                         & \checkmark &                            & \checkmark                  &                            &                         &                          &                         &                         & \checkmark & $1712\pm50$                    & -                             & 175         \\
              & \checkmark               & \checkmark               &           &                            & \checkmark                  &                            &                         &                          &                         &                         & \checkmark & $1649\pm50$                    & -                             & 182         \\ \hline
Loss          & \checkmark               & \checkmark               &           & \checkmark                  & \checkmark                  & \checkmark                  &                         &                          &                         &                         & \checkmark & \textbf{1148$\pm$50}          & -                             & 145         \\
              & \checkmark               & \checkmark               &           &                            & \checkmark                  & \checkmark                  &                         &                          &                         &                         & \checkmark & $1185\pm50$                    & -                             & 126         \\
              & \checkmark               & \checkmark               &           &                            &                            & \checkmark                  &                         &                          &                         &                         & \checkmark & $2008\pm50$                    & -                             & 123         \\ \hline
Classifier    & \checkmark               & \checkmark               &           &                            & \checkmark                  &                            & \checkmark               &                          &                         &                         &           & $1384\pm50$                    & $11.5\pm5.3$                    & -           \\
              & \checkmark               & \checkmark               &           &                            & \checkmark                  &                            &                         & \checkmark                &                         &                         &           & $1384\pm50$                    & $52.8\pm9.4$                    & -           \\
              & \checkmark               & \checkmark               &           &                            & \checkmark                  &                            &                         &                          & \checkmark               &                         &           & $1384\pm50$                    & $78.9\pm3.5$                    & -           \\
              & \checkmark               & \checkmark               &           &                            & \checkmark                  &                            &                         &                          &                         & \checkmark               &           & $1384\pm50$                    & $79.4\pm5.7$                    & -           \\ \hline
\end{tabular}
}
\end{small}
\end{table*}
\par
During the training process described in TABLE \ref{tab:Component discussion}, 2000 epochs are utilized to achieve convergence with a batch size of 4. The NVIDIA 1650 Ti GPU is employed for running, testing, and recording the following modules. It should be noted that when the batch normalization layer is not selected, the gradient explosion may occur, rendering all the previous training processes invalid.
\
\subsection{Implementation details}
\label{Implementation}
\begin{table}[H]
 
\label{tab:implement}
\caption{hyper-parameters}
\label{tab:hyper-parameters}
\centering
\setlength{\tabcolsep}{0.3mm}{
\begin{tabular}{c|ccc}
\hline
Hyper-param           & \multicolumn{3}{c}{Process}                                                                                                                                                     \\ \hline
name                  & \multicolumn{1}{c|}{\begin{tabular}[c]{@{}c@{}}coder\\ training\end{tabular}} & \multicolumn{1}{c|}{\begin{tabular}[c]{@{}c@{}}matrix\\ training\end{tabular}} & classification \\ \hline
batch size            & \multicolumn{1}{c|}{4}                                                        & \multicolumn{1}{c|}{1}                                                         & -              \\ \hline
optimizer             & \multicolumn{1}{c|}{Adam}                                                     & \multicolumn{1}{c|}{Adam}                                                      & -              \\ \hline
learning rate         & \multicolumn{1}{c|}{0.001}                                                    & \multicolumn{1}{c|}{0.010}                                                     & -              \\ \hline
epochs                & \multicolumn{1}{c|}{2000}                                                     & \multicolumn{1}{c|}{400}                                                       & -              \\ \hline
$K$ init mean         & \multicolumn{1}{c|}{1.000}                                                    & \multicolumn{1}{c|}{-}                                                         & -              \\ \hline
$K$ init variance     & \multicolumn{1}{c|}{4.000}                                                    & \multicolumn{1}{c|}{-}                                                         & -              \\ \hline
resolution $w$        & \multicolumn{2}{c|}{$64*64$}                                                                                                                                   & -              \\ \hline
cycle length $T$     & \multicolumn{2}{c|}{12}                                                                                                                                        & -              \\ \hline
matrix $K$ size       & \multicolumn{2}{c|}{$64*64$}                                                                                                                                   & -              \\ \hline
FC units $f$          & \multicolumn{2}{c|}{2048}                                                                                                                                      & -              \\ \hline
FC units $g$          & \multicolumn{2}{c|}{2048}                                                                                                                                      & -              \\ \hline
epoch                 & \multicolumn{2}{c|}{-}                                                                                                                                         & 2000           \\ \hline
regularization weight & \multicolumn{2}{c|}{-}                                                                                                                                         & 200\\      \hline     
\end{tabular}
}
\end{table}
\par
We defined the Koopman operator as a square matrix with the same shape as the video frame. Therefore, the shape of frames in the embedding space remains unchanged after continuous multiplication. To retain enough details while avoiding too many nodes, we defined the input frame shape as $[64, 64]$. In the optimal video segment (OVS) process, a gait cycle comprising 12 consecutive images was found to be appropriate. Thus, the shape of the three-dimensional data of a gait cycle is $[12, 64, 64]$. For each equally-transform (ET) process, we used a dense layer with $w*0.5w=2048$ units. During classification, the logistic model's regularization weight parameter was set to 200, and the max-iteration was set to 2000. All hyper-parameter settings are shown in TABLE \ref{tab:hyper-parameters}.

\section{Conclusion}
InvKA model employs an invertible autoencoder to map the periodic gait data stream into the low-dimensional embedding space, ensuring no loss during the training process. This bijective mapping enables the Koopman operator to represent the kinematic features of gait cycles, as shown in Fig. \ref{fig:K}. The eigenvalues of the Koopman operator correspond to angular velocities or expansion/shrinkage rates, providing better interpretability and making it more suitable for highly sensitive scenarios. Furthermore, not using convolution-deconvolution image autoencoders significantly reduces computational costs, enabling large-scale deployment of gait recognition technology. 
\newpage
\bibliographystyle{IEEEtran}
\bibliography{bare_jrnl}

\newpage
\appendices
\section{Synthetic frames generation}
\label{forgery}
For a gait video $\mathbb{G}:g^{w*w}_{t+1}=f_{t}g^{w*w}_{t}$, suppose the the representation of gait video in embedding space is $ \phi(X_{t}) $, since $\phi(X_{t})=K \phi(X_{t-1})=K^{2} \phi(X_{t-2})=\cdots=K^{t} \phi(X_{0})$. 
\par
\begin{figure}[htb]
    \begin{center}
        \includegraphics[width=0.4\textwidth]{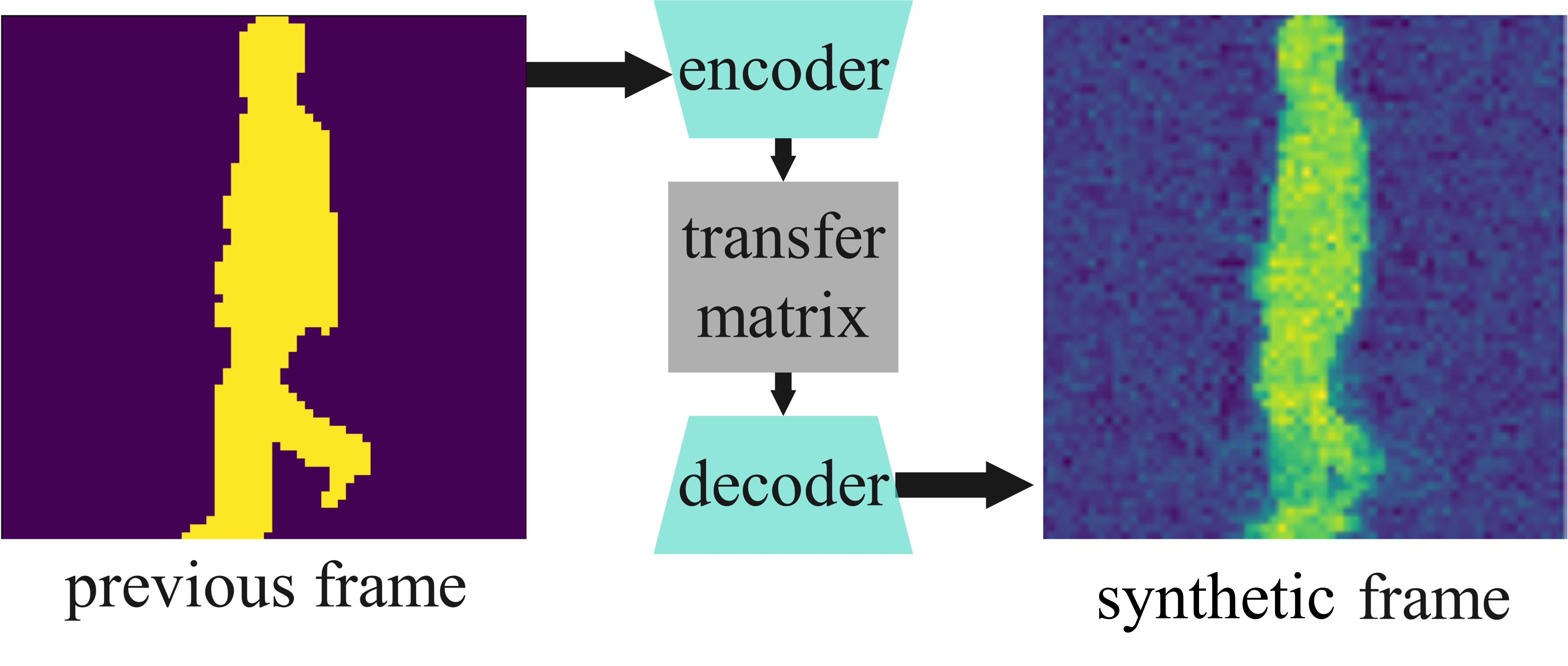}
        \caption{\label{fig:generateFake}InvKA model can generate synthesized images at other time points through the decoder’s output.}
    \end{center}
\end{figure}

 
Therefore, for the human motion in the time $ t+m $, the representation in embedding space can be easily calculated by $ \phi(X_{t+m})=K^{t+m} \phi(X_{0})$. the human motion of time $t+m$ in the Spatio-temporal space can be calculated using the invertible decoder. Fig. \ref{fig:generateFake} is the process of synthetic gait frame generation, the left figure represents the original frame $G_t$, and the right figure represents the synthetic frame $\phi^{-1}(K\phi(G))$.
\par
\begin{figure*}[!htb]
    \begin{center}
        \includegraphics[width=1\textwidth]{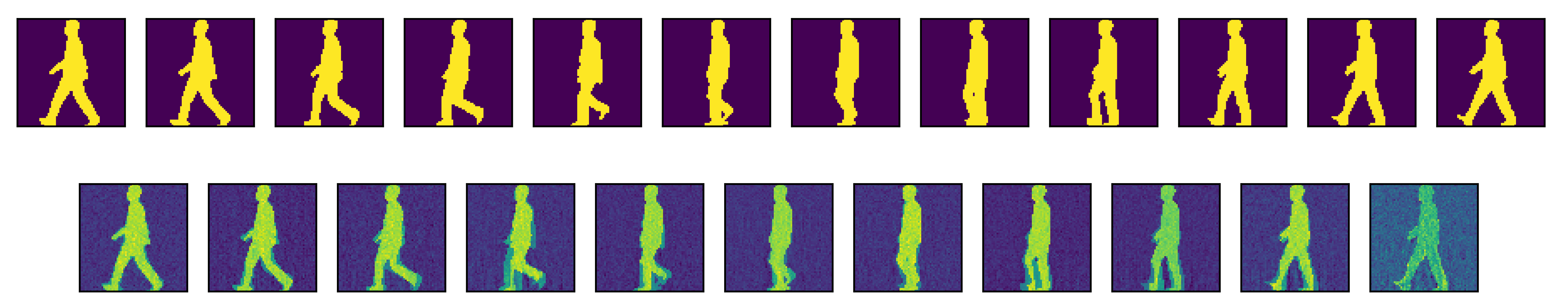}
        \caption{\label{fig:interpolation}original frames and synthetic frames, each frame in the second row is an synthetic interpolated image of adjacent frames in the first row.}
    \end{center}
\end{figure*}

 
The frames between two video frames can also be calculated with this method. For instance, the frames the $ n-th $ frame and the $ n+1-th $ frame can be calculated with $ \phi(X_{r})=K^{r} \phi(X_{0})$, where $ r $ is the corresponding real number between $n$ and $n+1$. Fig. \ref{fig:interpolation} is the experiment to insert a frame between two frames. Each frame in the second row is an interpolated image of adjacent frames in the first row. Therefore, it is easy to implement high frame rate video reconstruction with the proposed method.
\par
\begin{figure}[!htb]
    \begin{center}
        \includegraphics[width=0.48\textwidth]{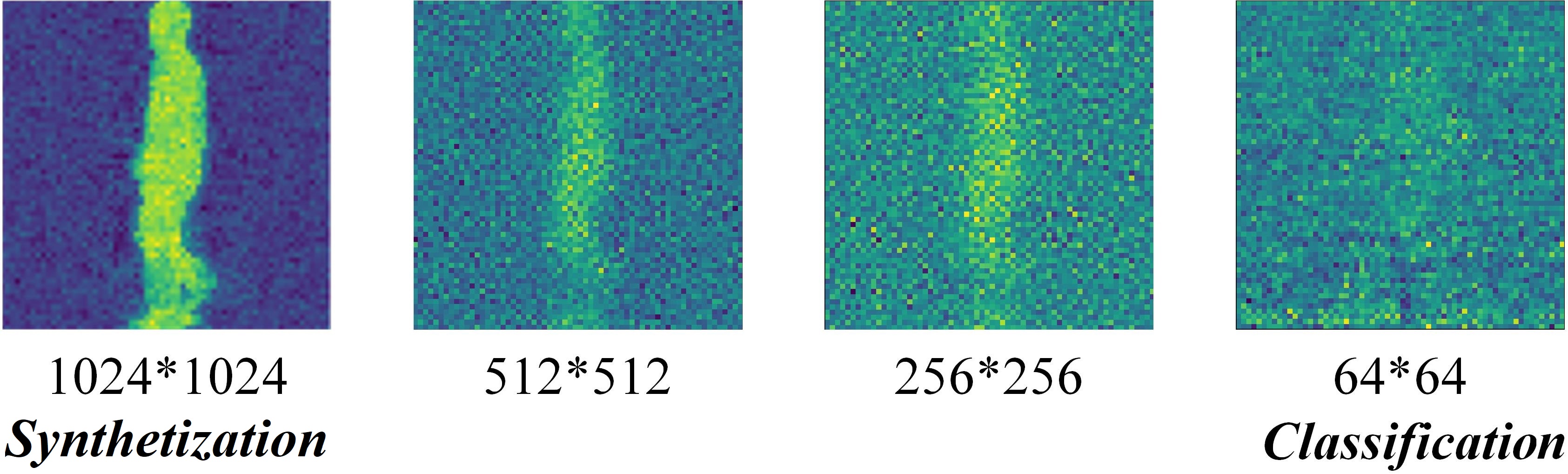}
        \caption{\label{fig:resolution}synthetic gait frames generated using Koopman operator with widths of 1024, 512, 256, and 64.}
    \end{center}
\end{figure}

It is worth noting that the autoencoder used for generating forged motion images has a higher number of nodes in the fully connected layer, i.e., higher Koopman operator size. The model used for classification and the model used for synthetic frame generation are not exactly the same. As Fig. \ref{fig:resolution} shows, fake images generated using Koopman operator with widths of 1024, 512, 256, and 64 has different quality. The larger the amount of information accommodated by the Koopman operator, the higher the resolution of the fake image generated. A width of 1024 or higher is the matrix size used in synthetization tasks, while a width of 64 is used in the gait classification process. TABLE \ref{similarity compare} employs various similarity metrics to measure the likeness between synthetic and real images. 

\begin{center}

\begin{table}[htb]
\centering
\caption{similarity scores on different node numbers}
\label{similarity compare}
\centering
\begin{tabular}{cc|cccc}
\hline
\multicolumn{2}{c|}{Simularity}          & \multicolumn{4}{c}{ET module nodes} \\ \hline
\multicolumn{1}{c|}{Index}     & Scale   & 1024    & 512     & 256    & 64     \\ \hline
\multicolumn{1}{c|}{$MSE_{sim}$} & (0,1)   & 0.86    & 0.72    & 0.67   & 0.60   \\ \hline
\multicolumn{1}{c|}{PSNR}      & (0,$\inf$) & 17.07  & 11.05  & 9.62 & 7.96 \\ \hline
\multicolumn{1}{c|}{UQI}       & (-1,1)  & 0.90    & 0.42    & 0.05   & $10^{-5}$   \\ \hline
\end{tabular}
\end{table}
\end{center}
\par

\begin{itemize}
    \item $MSE_{sim}$ (Mean Squared Error similarity) indicates the fundamental similarity between images.
    \item PSNR (Peak Signal Noise Ratio) evaluates the amount of noise in an image. A PSNR below 20 suggests significant noise in unfiltered synthetic gait data, likely caused by random gradient descent.
    \item UQI (Universal Quality Image Index) reflects the human eye’s perception of image similarity, with values closer to 1 or -1 indicating higher simulation accuracy. When the number of nodes decreases, the UQI index decreases significantly.
\end{itemize}
\par
InvKA demonstrates \textbf{controllable performance} in synthesising images, with a higher number of ET module nodes resulting in greater simulation quality. However, synthesized gait frames may contain numerous noise points, necessitating the use of a filter for denoising.
\section{Limitation}
\label{limitation}

\begin{table*}[htb]\centering
\caption{recognition rank-1 accuracy (\%) cross views under BG and CL on CASIA-B Dataset($\%$)}
\label{tab:robust acc}
\centering
 
\setlength{\tabcolsep}{0.5mm}{
\begin{tabular}{ccc|cccccc|c|cccccc|c}
\hline
\multicolumn{3}{c|}{Gallery NM$\#$1-4}                                 & \multicolumn{6}{c|}{BG $0^\circ$-$90^\circ$}                                                                   & Average           & \multicolumn{6}{c|}{CL $0^\circ$-$90^\circ$}                                                          & Average           \\ \hline
\multicolumn{3}{c|}{Probe View}                                        & $0^\circ$                      & $18^\circ$    & $36^\circ$    & $54^\circ$    & $72^\circ$    & $90^\circ$    & $0^\circ$-$90^\circ$ & $0^\circ$       & $18^\circ$      & $36^\circ$    & $54^\circ$      & $72^\circ$    & $90^\circ$      & $0^\circ$-$90^\circ$ \\ \hline
\multicolumn{1}{c|}{ST(24)} & \multicolumn{1}{c|}{Method}       & Year &                                &               &               &               &               &               &                      &                 &                 &               &                 &               &                 &                      \\
\multicolumn{1}{c|}{}       & \multicolumn{1}{c|}{MT3DCNN}      & 2020 & 64.5                           & 76.7          & 82.8          & 82.8          & 73.2          & 66.9          & 74.48                & 46.6            & 61.6            & 66.5          & 57.4            & 52.1          & \textbf{58.1}   & 57.05                \\
\multicolumn{1}{c|}{}       & \multicolumn{1}{c|}{GaitGL }       & 2021 & \textbf{68.1} & \textbf{81.2} & \textbf{87.7} & \textbf{84.9} & 76.3          & 70.5          & \textbf{78.12}       & 46.9            & 58.7            & \textbf{66.6} & 65.4            & \textbf{58.3} & 54.1            & \textbf{58.33}       \\
\multicolumn{1}{c|}{}       & \multicolumn{1}{c|}{GaitSet }      & 2022 & 64.1                           & 76.4          & 81.4          & 82.4          & \textbf{77.2} & \textbf{71.8} & 75.55                & 36.4            & 49.7            & 54.6          & 49.7            & 48.7          & 45.2            & 47.38                \\
\multicolumn{1}{c|}{}       & \multicolumn{1}{c|}{\textbf{InvKA}}  &      & 56.2                           & 56.7          & 41.3          & 52.0          & 63.2          & 65.2          & 55.77                & $\textbf{63.8}$ & $\textbf{79.5}$ & 52.3          & $\textbf{85.2}$ & 34.7          & $\textbf{55.6}$ & $\textbf{61.85}$     \\ \hline
\multicolumn{1}{c|}{MT(62)} & \multicolumn{1}{c|}{AE }           & 2017 & 29.8                           & 37.7          & 39.2          & 40.5          & 43.8          & 37.5          & 38.08                & 18.7            & 21.0            & 25.0          & 25.1            & 25.0          & 26.3            & 23.52                \\
\multicolumn{1}{c|}{}       & \multicolumn{1}{c|}{MGAN }         & 2018 & 48.5                           & 58.5          & 59.7          & 68.0          & 53.7          & 49.8          & 56.37                & 56.1            & 34.5            & 36.3          & 33.3            & 32.9          & 32.7            & 37.63                \\
\multicolumn{1}{c|}{}       & \multicolumn{1}{c|}{MT3DCNN }      & 2020 & 86.7                           & 92.9          & 94.9          & 92.8          & 88.5          & 82.5          & 89.72                & 67.5            & 81.0            & 85.0          & 80.6            & 75.9          & 69.8            & 76.63                \\
\multicolumn{1}{c|}{}       & \multicolumn{1}{c|}{GaitGL }       & 2021 & \textbf{88.5}                  & \textbf{95.1} & \textbf{95.9} & \textbf{94.2} & \textbf{91.5} & \textbf{85.4} & \textbf{91.77}       & \textbf{70.7}   & \textbf{83.2}   & \textbf{87.1} & \textbf{84.7}   & \textbf{78.2} & \textbf{71.3}   & 79.20                \\
\multicolumn{1}{c|}{}       & \multicolumn{1}{c|}{GaitSet }      & 2022 & 79.9                           & 89.8          & 91.2          & 86.7          & 81.6          & 76.7          & 84.32                & 52.0            & 66.0            & 72.8          & 69.3            & 63.1          & 61.2            & 64.07                \\
\multicolumn{1}{c|}{}       & \multicolumn{1}{c|}{\textbf{InvKA}}  &      & 55.1                           & 47.2          & 31.5          & 50.2          & 50.3          & 64.3          & 49.77                & 62.9            & 76.6            & 38.4          & $\textbf{80.6}$ & 55.3          & 59.3            & 62.18                \\ \hline
\multicolumn{1}{c|}{LT(74)} & \multicolumn{1}{c|}{CNN-LB }       & 2016 & 64.2                           & 80.6          & 82.7          & 76.9          & 64.8          & 63.1          & 72.05                & 37.7            & 57.2            & 66.6          & 61.1            & 55.2          & 54.6            & 55.40                \\
\multicolumn{1}{c|}{}       & \multicolumn{1}{c|}{Gait-RNNPart } & 2020 & 86.0                           & 93.3          & 95.1          & 92.1          & 88.0          & 82.3          & 89.47                & 65.8            & 80.7            & 82.5          & 81.1            & 72.7          & 71.5            & 75.72                \\
\multicolumn{1}{c|}{}       & \multicolumn{1}{c|}{MT3DCNN }      & 2020 & 91.0                           & 95.4          & 97.5          & 94.2          & 92.3          & 86.9          & 92.88                & 76.0            & 87.6            & 89.8          & 85.0            & 81.2          & 75.5            & 82.52                \\
\multicolumn{1}{c|}{}       & \multicolumn{1}{c|}{GLN }          & 2020 & 91.0                           & \textbf{97.7} & 97.8          & 95.2          & 92.5          & 91.2          & 94.23                & 70.6            & 82.4            & 85.2          & 82.7            & 79.2          & 76.4            & 79.42                \\
\multicolumn{1}{c|}{}       & \multicolumn{1}{c|}{GaitPart }     & 2020 & 89.1                           & 94.8          & 96.7          & 95.1          & 88.3          & \textbf{94.9} & 93.15                & 70.7            & 85.5            & 86.9          & 83.3            & 77.1          & 72.5            & 79.33                \\
\multicolumn{1}{c|}{}       & \multicolumn{1}{c|}{GaitGL }       & 2021 & 92.6                           & 96.6          & 96.8          & 95.5          & 93.5          & 89.3          & 94.05                & 76.6            & 90.0            & 90.3          & 87.1            & 84.5          & 79.0            & 84.58                \\
\multicolumn{1}{c|}{}       & \multicolumn{1}{c|}{3dLocalMT }    & 2021 & 92.9                           & 95.9          & 97.8          & \textbf{96.2} & 93.0          & 87.8          & 93.93                & 78.2            & 90.2            & 92.0         & 87.1            & 83.0          & 76.8            & 84.55                \\
\multicolumn{1}{c|}{}       & \multicolumn{1}{c|}{CSTL }         & 2021 & \textbf{95.0}                  & 96.8          & \textbf{97.9} & 96.0          & \textbf{94.0} & 90.5          & \textbf{95.03}       & \textbf{84.1}   & 92.1            & 91.8          & 87.2            & 84.4          & 81.5            & \textbf{86.85}                \\
\multicolumn{1}{c|}{}       & \multicolumn{1}{c|}{RPNet }        & 2022 & 92.3                           & 96.6          & 96.6          & 94.5          & 91.9          & 87.6          & 93.25                & 75.6            & 87.1            & 88.3          & 83.1            & 78.8          & 78.0            & 81.82                \\
\multicolumn{1}{c|}{}       & \multicolumn{1}{c|}{GaitSet }      & 2022 & 86.7                           & 94.2          & 95.7          & 93.4          & 88.9          & 85.5          & 90.73                & 59.5            & 75.0            & 78.3          & 74.6            & 71.4          & 71.3            & 71.68                \\
\multicolumn{1}{c|}{}       & \multicolumn{1}{c|}{GaitNet }      & 2022 & 83.0                           & -             & -             & 86.6          & -             & 74.8          & 81.47                & 42.1            & -               & -             & 70.7            & -             & 70.6            & 61.13                \\
\multicolumn{1}{c|}{}       & \multicolumn{1}{c|}{DyGait }       & 2023 & 94.5                           & 96.9          & 97.4          & 96.1          & 95.4          & 94.0          & 95.72                & 82.2            & \textbf{93.0}            & \textbf{95.2}          & \textbf{91.6}            & \textbf{87.1}          & \textbf{83.4}            & 88.75                \\
\multicolumn{1}{c|}{}       & \multicolumn{1}{c|}{GLN-HBS }      & 2023 & 91.7                           & 96.6          & 97.3          & 95.9          & 93.7          & 89.5          & 94.12                & 77.7            & 89.4            & 91.9          & 87.0            & 84.1          & 78.1            & 84.70                \\
\multicolumn{1}{c|}{}       & \multicolumn{1}{c|}{GaitSet-HBS }  & 2023 & 89.7                           & 93.8          & 94.6          & 92.9          & 88.2          & 83.0          & 90.37                & 72.9            & 84.1            & 83.7          & 79.6            & 73.0          & 70.5            & 77.30                \\
\multicolumn{1}{c|}{}       & \multicolumn{1}{c|}{\textbf{InvKA}}  &      & 54.9                           & 45.6          & 32.4          & 49.1          & 49.7          & 61.2          & 48.82                & 62.7            & 71.8            & 37.5          & $\textbf{81.0}$ & 39.7          & 61.1            & 58.97                \\ \hline
\end{tabular}}

\end{table*}
\par
\textbf{\emph{Robustness}}: By evaluating the performance of the model on the BG (bag) and Cl (clothes) datasets, TABLE \ref{tab:robust acc} demonstrates that the model exhibits poor robustness to other modal data with interference. This may be due to the fact that the model is composed of several separate modules, which can lead to information loss during the cutting and segmenting of the dataset. Clothing and backpacks, which have poor stability and may have different cycle lengths compared to the human body, often obscure the details of people's motions when calculating the kinematic feature matrix. The loss of information related to these types of motions leads to a decrease in the quality of the trained coder, resulting in less representative estimates of the kinematic feature matrix. However, InvKA model still get competitive precision on certain views, especially on small-scale datasets.

\par\textbf{\emph{Possibility of under fitting}}:
\label{comparison}
Our model, similar to the OFA model \cite{zhang2021cross}, is based on Koopman operator theory and uses neural networks as coders/decoders. Unfortunately, the OFA model utilizing the convolution-deconvolution structure did not report recognition accuracy on the CASIA-B dataset. TABLE \ref{tab:OFACompare} compares the accuracy of our model and the OFA series models on the OU-MVLP dataset. 
\begin{table}[htb]
\centering
\caption{rank-1 accuracy comparison on OU-MVLP dataset}
\label{tab:OFACompare}
\setlength{\tabcolsep}{2mm}{
\begin{tabular}{l|llll}
\hline
Angle        & $0^\circ$    & $30^\circ$   & $60^\circ$   & $90^\circ$   \\ \hline
OFA+KMMD+DFE & 44.9 & 80.5 & 88.3 & 95.4 \\ \hline
\textbf{InvKA(Ours)}         & 67.1 & 65.8 & 60.2 & 55.4 \\ \hline
\end{tabular}
}
\end{table}
The OU-MVLP dataset\cite{takemura2018multi} is a cross-view dataset with only \textbf{2 sequences} provided for each angle. This results in the OFA module producing a significantly smaller number of accurately segmented gait cycles per trial compared to the CASIA-B dataset, which provides \textbf{6 sequences} for each angle. As logistic regression classifiers are sensitive to the data scale(See TABLE \ref{tab:sampleCount}), a small training set results in suboptimal performance due to the under-fitting. Another possible factor is that the autoencoder has a concise structure, which may limit its robustness to occluded frames(See  Section \ref{Robustness testing}). However, as Fig. \ref{fig:performance} shows, our model has \textbf{significantly lower computational cost} compared with OFA model.


\begin{table}[htb]
\centering
\caption{cross-validation rank-1 accuracy($\%$) on different data scale}
\setlength{\tabcolsep}{1.5mm}{
\label{tab:sampleCount}
\begin{tabular}{c|ccccccc}

\hline 
$Scale_{avg}$ & 5.8  & 7.9   & 12.0 & 16.1 & 20.2 & 22.0 & 24.7 \\ \hline
ST(24)           & 27.0 & 41.7  & 48.4 & 58.9 & 66.0 & 63.5 & 89.7 \\ \hline
MT(62)           & 28.6 & 37.38 & 45.4 & 57.2 & 63.3 & 59.0 & 86.4 \\ \hline
LT(74)           & 27.8 & 35.6  & 43.4 & 55.4 & 61.7 & 58.9 & 83.9 \\ \hline
\end{tabular}}
\end{table}
\par
\label{Robustness testing}

\section{Proof for convexity}
\label{proof}The gait cycle tensor with shape $[T,W,W]$ is represented by $\textbf{G}$.
%
%
\par
From (\ref{Eq:phi}), $\textbf{X}$ is the tensor converted by $\phi(\textbf{G})$ with same shape $[T,W,W]$. By using the chain rule:

\begin{equation} 
\label{Eq:partial2}
\begin{aligned}
		   \frac{\partial L(\textbf{G},\phi, K)}{\partial K} &=\frac{\partial L(\textbf{X}=\phi(\textbf{G}), K)}{\partial K} \\
     &= \sum_{t=1}^{T}\frac{\partial L(X_t,X_{t+1}, K)}{\partial K} \\
     &= \sum_{t=1}^{T}\frac{\partial L(X_t,X_{t+1}, K)}{\partial X_{t+1}}\frac{\partial X_{t+1}}{\partial K}\\
\end{aligned}
\end{equation}

From (\ref{Eq:phiXt}), the partial derivative of $X_{t+1}$ to $X_t$ is $K$. Thus, (\ref{Eq:partial2}) can be expressed as:

\begin{equation} \label{Eq:partial4}
		\begin{aligned}
		    \frac{\partial L(\textbf{G},\phi, K)}{\partial K} &= \sum_{t=1}^{T}(X_{t+1}-X_{t})X_t \\
                &= \sum_{t=1}^{T}X_{t+1}X_t-X_{t}^2
		\end{aligned}
\end{equation}

In order to judge concavity and convexity, the second partial derivative must be calculated. With formula (\ref{Eq:partial4}):

\begin{equation} \label{Eq:secpartial1}
	\begin{split}
		\begin{aligned}
		    \frac{\partial^2 L(\textbf{G},\phi, K)}{\partial K^2} &= \frac{\partial }{\partial K}\frac{\partial L(\textbf{X}, K)}{\partial K}\\ &= \sum_{t=1}^{T}\frac{\partial }{\partial K}(X_{t+1}X_t-X_t^2)
		\end{aligned}
	\end{split}
\end{equation}

By using chain rule on (\ref{Eq:secpartial1}), a conclusion can be made:

\begin{equation} \label{Eq:secpartial2}
	\begin{split}
		\begin{aligned}
		    \frac{\partial^2 L(\textbf{G},\phi, K)}{\partial K^2} &= \sum_{t=1}^{T}\frac{\partial }{\partial X_{t+1}}(X_{t+1}X_t-X_t^2)\frac{\partial X_{t+1}}{\partial K} \\
      &=  \sum_{t=1}^{T}X_t^2 = \textbf{X}^2= \phi(\textbf{G})^2\geq 0\\
		\end{aligned}
	\end{split}
\end{equation}
\par Thus, the loss function $L$ is a convex function with respect to $K$.
\par
\section{Proof for least-squares method}
\label{sec:LS}
After completing the autoencoder training process, the model can choose not to rely on the neural network to approximate $K$ but to obtain the analytic solution of $K$ directly through a tedious least squares process. The complete mathematical derivation of this method is interpretable. However, due to computational efficiency considerations, the approximate solution is used in the experimental data. The algorithm of least squares for analytic solutions can provide a better interpretation of this model.
\par
To simplify the derivation process and the symbol system. In the next derivation of the least squares algorithm, with  Eq. (\ref{Eq:phi}) and  Eq. (\ref{Eq:loss1}) we define.
\begin{equation}
\textbf{Y}_{true}^i := \textbf{X}^{i+1}\\
\end{equation}
\begin{equation} \label{Eq:Y_d}
	\textbf{Y}_{pred}^i := K\textbf{X}^i
\end{equation}
To simplify the derivation process, we define $J$:
\begin{equation}
\begin{aligned}
    J &= \frac{1}{2} Loss_1\\
    &= \frac{1}{2}\sum(\textbf{Y}_{true}-\textbf{Y}_{pred})^2
\end{aligned}
\end{equation}
\par
Both $\textbf{X}$ and $\textbf{Y}$ are tensors of shape [L, W, W]. L and W are hyperparameters. By the Koopman operator theory, the model should have the ability to predict $\textbf{Y}$ by using $\textbf{X}$ and $K$, the prediction $\textbf{Y}^i$ can be written as the matrix product of $K$ and $\textbf{X}^i$, where $K$ is the Koopman operator. 

$\textbf{X}^i$ and $\textbf{Y}^i$ are each item in the tensor $\textbf{X}$ and $\textbf{Y}$ with shape [W, W]. The loss value can be considered as the sum of the interpolated squares of each element in the $\textbf{X}^i$ and $\textbf{Y}^i$ matrices.
So in this problem, the loss function J can be written as:
\begin{equation} \label{Eq:Loss_d}
	J = \frac{1}{2}\sum_{i=1}^{L}\sum_{j=1}^{W}\sum_{k=1}^{W}(\textbf{Y}_{true}^{i,j,k}-\textbf{Y}_{pred}^{i,j,k})^2
\end{equation}

From  Eq. (\ref{Eq:Loss_d}) and  Eq. (\ref{Eq:Y_d}):
\begin{equation} \label{Eq:J}
		J = \frac{1}{2}\sum_{i=1}^{T}\sum^{W}\sum^{W}(\textbf{Y}_{true}^{i}-K\textbf{X}^i)^T(\textbf{Y}_{true}^{i}-K\textbf{X}^i)
\end{equation}
For convenience, we use L as the loss value of each item in J. $\textbf{Y}_{true}$ is written as $\textbf{Y}$:
\begin{equation} \label{Eq:L}
	L = \frac{1}{2}\sum^{W}\sum^{W}(\textbf{Y}-K\textbf{X})^T(\textbf{Y}-K\textbf{X})
\end{equation}
$K$, $X$ and $Y$ can be written as:
\begin{equation} 
K=
\begin{bmatrix}
k_1^1 & \cdots &  k_1^W\\
\vdots & \ddots &  \vdots\\
k_W^1 & \cdots & k_W^W \\
\end{bmatrix}
\end{equation}
\begin{equation} 
X=
\begin{bmatrix}
x_1^1 & \cdots &  x_1^W\\
\vdots & \ddots &  \vdots\\
x_W^1 & \cdots & x_W^W \\
\end{bmatrix}
\end{equation}
\begin{equation} 
Y=
\begin{bmatrix}
y_1^1 & \cdots &  y_1^W\\
\vdots & \ddots &  \vdots\\
y_W^1 & \cdots & y_W^W \\
\end{bmatrix}
\end{equation}

Thus, L can be written as:

\begin{equation} \label{Eq:compute}
\begin{aligned}
	L =\frac{1}{2}\sum_{i}^W\sum_{j}^W(
y_i^j-\sum_n^Wk_i^nx_n^j
)^2
\end{aligned}
\end{equation}

Then do the partial derivative of L with respect to k:

\begin{equation} \label{Eq:derivative}
    \frac{\partial L}{\partial k_n^m} = \sum_i^W k_n^i\left(
    \begin{bmatrix}
    x_m^1\\
    \vdots\\
    x_m^W\\
    \end{bmatrix}^T
    \begin{bmatrix}
    x_i^1\\
    \vdots\\
    x_i^W\\
    \end{bmatrix}
    \right)-
    \begin{bmatrix}
    x_m^1\\
    \vdots\\
    x_m^W\\
    \end{bmatrix}^T
    \begin{bmatrix}
    y_n^1\\
    \vdots\\
    y_n^W\\
    \end{bmatrix}
\end{equation}
Define $X$ as the row vector of $\textbf{X}$, $Y$ as the row vector of $\textbf{Y}$:
\begin{equation}\label{Eq:rowX}
    X_i := 
    \begin{bmatrix}
    x_i^1 & \cdots & x_i^W
    \end{bmatrix}
\end{equation}
\begin{equation}\label{Eq:rowY}
    Y_i := 
    \begin{bmatrix}
    y_i^1 & \cdots & y_i^W
    \end{bmatrix}
\end{equation}
From  Eq. (\ref{Eq:derivative}),  Eq. (\ref{Eq:rowX}) and  Eq. (\ref{Eq:rowY}):
\begin{equation} \label{Eq:derFin}
    \frac{\partial L}{\partial k_n^m} = \sum_i^W k_n^i(
    X_m^T
    X_i
    )-
    X_m^T
    y
\end{equation}
Set the partial derivative to 0:
\begin{equation} \label{Eq:poly}
\begin{aligned}
        \frac{\partial L}{\partial k_n^m} = \sum_i^W k_n^i(X_m^TX_i)- X_m^Ty &=0\\
        \sum_i^W k_n^i(X_m^TX_i) &= X_m^Ty
\end{aligned}
\end{equation}
Finding each term of K is solving a system of polynomial equations of Eq. (\ref{Eq:poly}). Thus, rewrite Eq. \eqref{Eq:poly} to the augmented matrix $A$. Adding up the corresponding elements of $A$ to each term in J gives the augmentation matrix $S$ of the whole problem:
\begin{equation}
    S = \sum_i^LA_i
\end{equation}

\begin{figure}[!htb]
  \centering
  \includegraphics[width=0.3\textwidth]{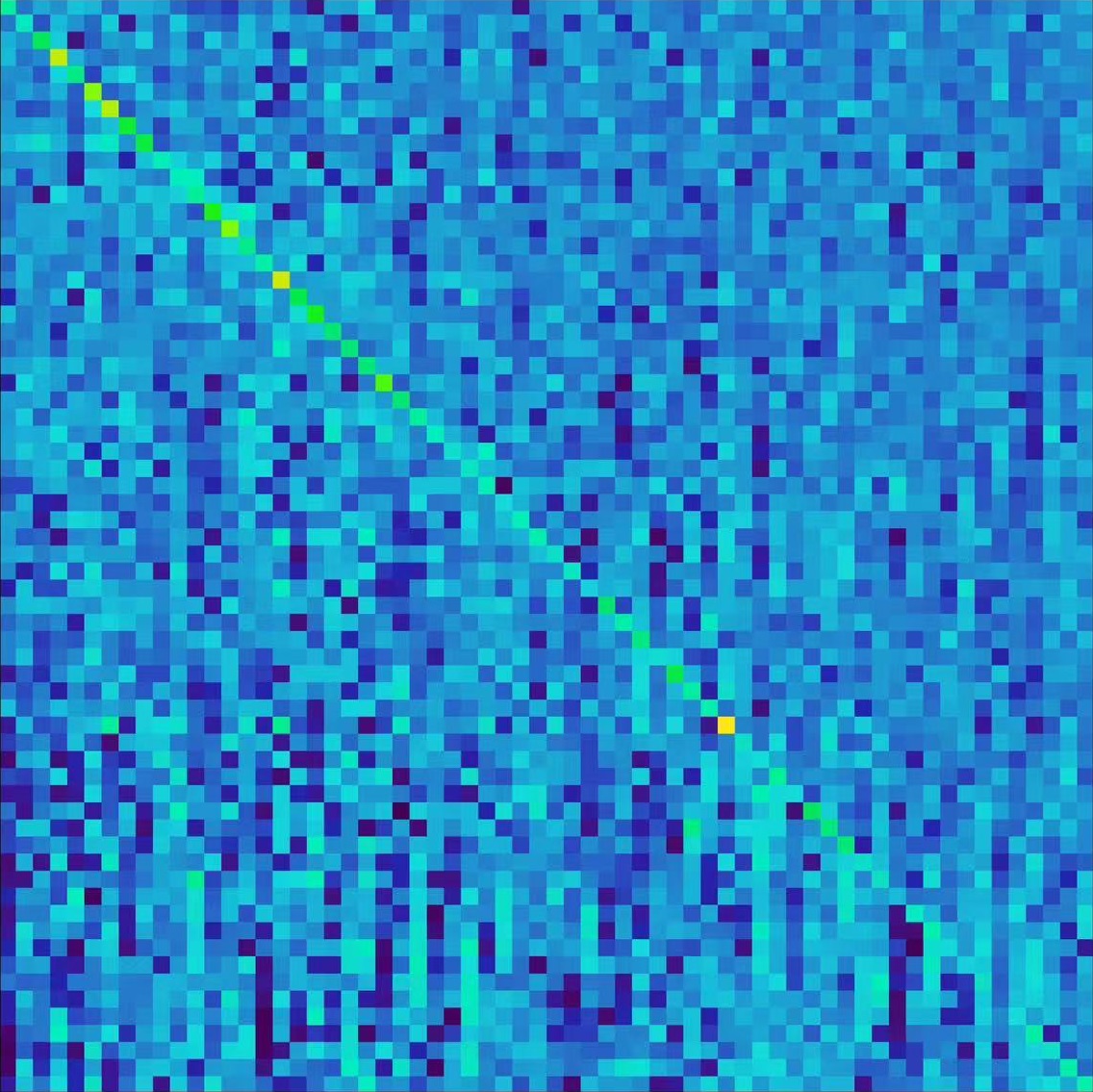}
  \caption{\label{fig:LS_val}Closed-format solution obtained by least squares approach.}
\end{figure}

As Fig. \ref{fig:LS_val} shows, the analytical solution obtained by least squares approach is similar to the approximate solution obtained by gradient descent, and the representative diagonal appears in both matrices.
\section{Weight matrices}
\label{coef}
\begin{figure}[!htb]
  \centering
  \includegraphics[width=0.5\textwidth]{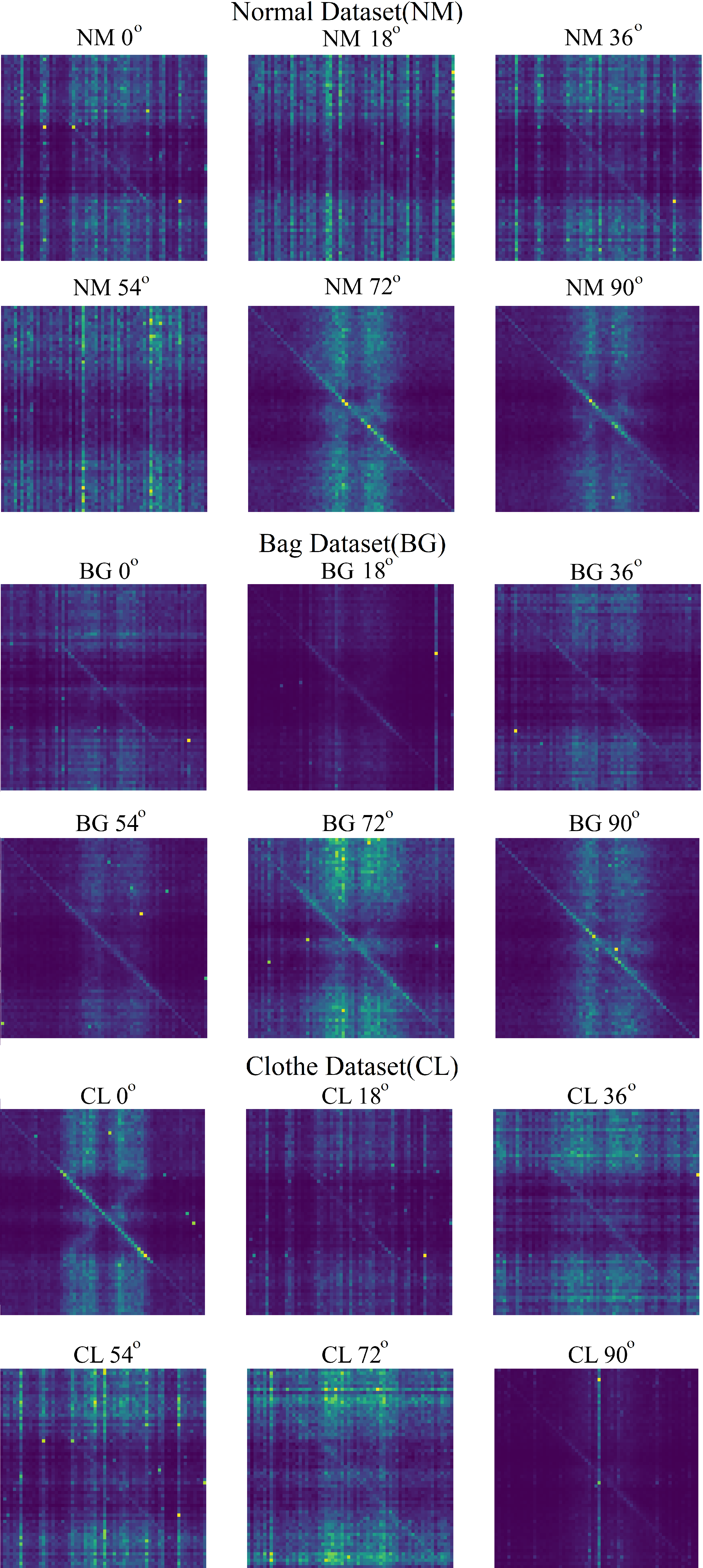}
  \caption{\label{fig:Coef_all}weight matrices of logistic regression classifier of all galleries.}
\end{figure}
In the NM series dataset, a deviation from a $90^\circ$ shooting angle leads to a significant increase in occlusion between human limbs and torso, resulting in a reduction of unique gait features between samples. Consequently, classification is no longer reliant on large movements (since the differences in movement amplitude under non-frontal angles are weakened), which is reflected in the decreased weight of diagonal pixels and increased weight of non-diagonal pixels. Ultimately, the variance matrix becomes more symmetrically distributed. The data presented in Fig. \ref{fig:Coef_all} exemplifies this phenomenon, thereby confirming the model's interpretability.
\par
In the BG and CL datasets, clothing and backpacks can significantly occlude or interfere with human posture. Consequently, the classifier tends to focus on details, resulting in a variance matrix with off-diagonal elements that are close in value to the diagonal elements, as seen in BG $0^\circ$, BG $36^\circ$, BG $72^\circ$, BG $90^\circ$, CL $0^\circ$, CL $36^\circ$, and CL $72^\circ$. However, due to the smaller size of the BG and CL datasets, they are more susceptible to the influence of outliers, leading to overfitting where the classifier wrongly uses a few outlier feature points as the basis for classification. Examples of this include BG18 and CL90, where the variance images exhibit extreme values at certain specific pixels, as shown in Fig. \ref{fig:Coef_all}.

\end{document}